\documentclass[sigconf, screen]{acmart}
\usepackage{multirow}
\usepackage{makecell}
\usepackage{booktabs}
\usepackage{caption}
\usepackage{wrapfig}
\usepackage[rightcaption]{sidecap}
\usepackage{bm}
\AtBeginDocument{%
  \providecommand\BibTeX{{%
    \normalfont B\kern-0.5em{\scshape i\kern-0.25em b}\kern-0.8em\TeX}}}

\setcopyright{acmcopyright}
\copyrightyear{2020}
\acmYear{2020}
\acmDOI{10.1145/1122445.1122456}

\acmConference[MM '20]{MM '20: 28th ACM International Conference
  on Multimedia}{Oct 12--16, 2020}{Seattle, United States}
\acmPrice{15.00}
\acmISBN{978-1-4503-XXXX-X/18/06}

\begin{document}

\title{Learning to Simulate Complex Scenes}

\author{Zhenfeng Xue}
\affiliation{%
  \institution{Zhejiang University}
  \streetaddress{No.38 Zheda Road}
  \city{Hangzhou}
  \country{China}
  \postcode{310027}
}
\email{zfxue0903@zju.edu.cn}

\author{Weijie Mao}
\authornote{Corresponding author}
\affiliation{%
  \institution{Zhejiang University}
  \streetaddress{No.38 Zheda Road}
  \city{Hangzhou}
  \country{China}
}
\email{wjmao@zju.edu.cn}

\author{Liang Zheng}
\affiliation{%
  \institution{Australian National University}
  \streetaddress{ACT 2601}
  \city{Canberra}
  \country{Australia}}
\email{liang.zheng@anu.edu.au}

\renewcommand{\shortauthors}{Xue, et al.}

\begin{abstract}
Data simulation engines like Unity are becoming an increasingly important data source that allows us to acquire ground truth labels conveniently. Moreover, we can flexibly edit the \emph{content} of an image in the engine, such as objects (position, orientation) and environments (illumination, occlusion). 
When using simulated data as training sets, its editable content can be leveraged to mimick the distribution of real-world data, and thus reduce the content difference between the synthetic and real domains. 
This paper explores content adaptation in the context of semantic segmentation, where the complex street scenes are fully synthesized using 19 classes of virtual objects from a first person driver perspective and controlled by 23 attributes.
To optimize the attribute values and obtain a training set of similar content to real-world data, we propose a scalable discretization-and-relaxation (SDR) approach. 
Under a reinforcement learning framework, we formulate attribute optimization as a random-to-optimized mapping problem using a neural network.
Our method has three characteristics. 
1) Instead of editing attributes of individual objects, we focus on global attributes that have large influence on the scene structure, such as object density and illumination. 
2) Attributes are quantized to discrete values, so as to reduce search space and training complexity.
3) Correlated attributes are jointly optimized in a group, so as to avoid meaningless scene structures and find better convergence points. 
Experiment shows our system can generate reasonable and useful scenes, from which we obtain promising real-world segmentation accuracy compared with existing synthetic training sets.
\end{abstract}




\maketitle

\section{Introduction}
Collecting and annotating large-scale datasets \cite{Imagenet2009,geiger2013vision,zheng2015scalable} consumes much time and manpower.
This is especially true for semantic segmentation, where high-quality annotation is reported to require 60 or 90 minutes for an image \cite{brostow2009semantic,cordts2016cityscapes}. For this problem, data synthesis through graphic engines \cite{Richter_2016_ECCV,gaidon2016virtual,ros2016synthia} has recently become a promising solution due to its convenience of acquiring ground truths at a large scale. 
This strategy also enables us to simulate corner cases that are not well covered by mainstream datasets. 
Besides, model testing in virtual environment \cite{dosovitskiy2017carla,ai2thor,Wu2018BuildingGA} is safe and economic.
The challenge lies in the domain gap between synthetic and real-world data that leads to performance drop.

Synthetic-to-real domain adaptation is a popular way to narrow the domain gap \cite{zou2018domain,tsai2018learning,li2018semantic,hoffman2017cycada,hoffman2016fcns}. 
These methods attempt to solve this problem from two aspects, \textit{i.e.}, appearance level and feature level. For the former, stylized synthetic images are generated to resemble those captured in real world \cite{hoffman2017cycada,li2018semantic,chen2019learning} using some GAN-based methods \cite{NIPS2014_5423,Zhu_2017_ICCV,Isola_2017_CVPR,huang2018munit}. For the latter, the feature distributions among two domains are aligned \cite{hoffman2016fcns,tsai2018learning,luo2019taking}.

Several latest works reveal that there exists an under-explored but important aspect that causes the domain gap - the content difference \cite{prakash2019structured,ruiz2018learning,kar2019metasim,yao2019simulating}. Here, content may refer to building density, vehicle occlusions, illumination, \emph{etc}, which is different from the style gap 
and can hardly be addressed by most style- or feature-level adaptation methods, and required to be manually solved within virtual environment \cite{geiger2013vision}. To reduce manual effort, an emerging feasible solution is learning-based simulation \cite{ruiz2018learning,kar2019metasim,yao2019simulating}. To align the content between two domains, this strategy updates attribute values based on supervision signals such as distribution difference \cite{Gretton2012,Heusel2017GANsTB,yao2019simulating,kar2019metasim} or task loss \cite{kar2019metasim,ruiz2018learning}. This optimization process differs significantly from the traditional gradient based ones, mainly because the system is non-differentiable. In fact, the rendering function of the graphic engine is not known and thus is non-differentiable. Moreover, when calculating the task loss, the task model should be trained until convergence, and this process is non-differentiable, either. The high system complexity poses critical challenges if we aim to generate complex environments such as the street scenes.

\begin{figure*}[t]
    \centering
    \includegraphics[scale=0.5]{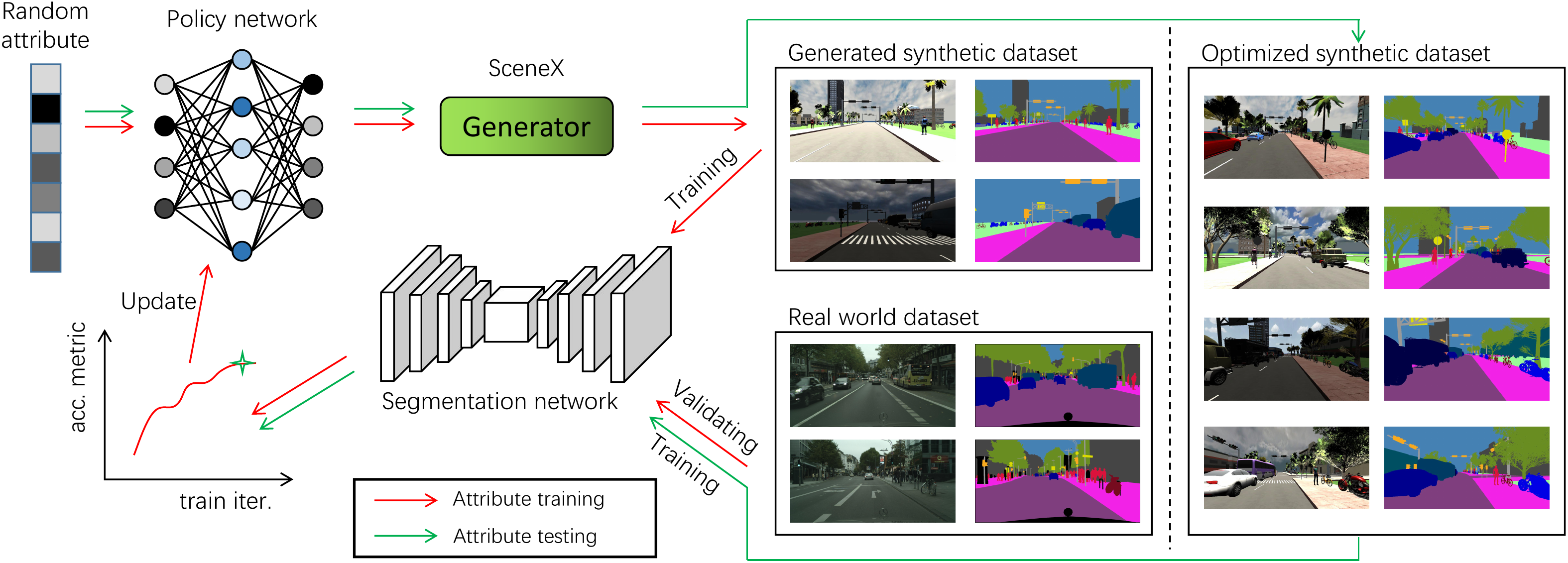}
    \caption{An overview of the proposed framework. During simulation learning, the attribute values are randomly initialized and fed into the policy network. The output of policy network is updated attribute values that are sent into the simulator (Unity) to render synthetic data. The synthetic data is used to train a segmentation model, and we use the model accuracy (mIoU) on real-world test set as reward to update the policy network. During inference, randomly initialized attributes are fed into the learned policy network, and the output attribute values are used to generate the optimized synthetic dataset.}
    \label{fig:overview}
\end{figure*}

We are thus interested in synthesizing large and complex scenes using the graphic engine and tackling the high computation problem with a relatively scalable approach. 
Existing approaches in this domain encounter difficulties when synthesizing large scenes, as compared in Table \ref{tab:method_difference}.
\textbf{First}, many methods optimize instance-level attributes, such as position and scale of each object \cite{dosovitskiy2017carla,prakash2019structured,tobin2017domain}. When a complex scene contains many objects, this practice will lead to a huge search space at the stage of scene structure optimization. Yao \emph{et al.} \cite{yao2019simulating} do not use instance-level attributes, but their method is designed for simple bounding boxes with a vehicle in the central and encounters efficiency problems under large scenes (Fig. \ref{fig:method_compare}). \textbf{Second}, in \cite{kar2019metasim,ruiz2018learning}, the search space for every attribute is continuous, which requires the REINFORCE algorithm to sample in a large range value. As the number of attribute increases, the search space becomes extremely large and the training complexity increases heavily. \textbf{Third}, the attributes are usually optimized sequentially \cite{kar2019metasim} or independently \cite{yao2019simulating}. These methods do not comprehensively consider the correlation among multiple attributes, and may cause object collisions under complex scenes (Fig. \ref{fig:generated_images}).

\begin{table}[t]
\caption{Differences with existing methods.}
    \centering
    \begin{tabular}{|p{1cm}|p{1.2cm}<{\centering}|p{1.4cm}<{\centering}|p{1.6cm}<{\centering}|p{1cm}<{\centering}|}
    \hline
         \multirow{2}{*}{Method}  & Attribute type  & Search space  & Attribute correlation  & Scene scale  \\
    \hline
         \cite{kar2019metasim}  & instance  & continuous  & partial  &medium \\ \hline
         \cite{ruiz2018learning} & instance & continuous  & partial  &medium \\ \hline
         \cite{yao2019simulating} & global & discrete  & none & simple \\ \hline
         Ours   & global & discrete & yes & complex \\ \hline
    \end{tabular}
    \label{tab:method_difference}
\end{table}

This paper proposes a scalable discretization-and-relaxation (SDR) data synthesis approach tailored for complex street scenes, so that a semantic segmentation model can be trained. An overview of the proposed framework is shown in Fig. \ref{fig:overview}. In a nutshell, our system uses a policy network to take proper actions to sample the optimal values of engine attributes, and the segmentation accuracy on real-world test sets serves as the system reward. This type of pipeline has also been adopted by existing works \cite{kar2019metasim,ruiz2018learning,yao2019simulating}. The distinct feature of our system consists of its scalability: efficient and effective optimization procedure is achieved. Specifically, our method addresses the three problems mentioned above. \textbf{First}, instead of using instance-level attributes, we build our system to accommodate global attributes, such as building density, lighting intensity, \emph{etc}. Our intuition is that global attributes would have large influence on the scene structure; moreover, there are much fewer global attributes compared with the large number of instance attributes, so we are facing a much smaller search space. \textbf{Then}, to reduce the search space caused by continuous attribute value sampling, SDR quantizes the attribute values into discrete values. To remedy the loss of randomness in attribute values, we add a relaxation step, \emph{i.e.,} we manually inject variance on top of the discrete values. \textbf{Finally}, the discretization process allows us to jointly optimize a group of attributes while maintaining a relatively low computational complexity. The joint optimization considers the correlation among attributes and can yield reasonable scene structures.

We perform the proposed optimization method on a new data synthesis platform named SceneX. It contains 19 classes of objects compatible with the mainstream semantic segmentation datasets, and to ensure diversity most classes have a rich range of 3D models. From SceneX, the proposed SDR method can generate a high-quality database where pixel-level annotations are accurately and automatically obtained. Comparing with existing synthetic datasets such as GTA5 and SYNTHIA that are manually designed, we show that optimized SceneX can yield very promising segmentation accuracy on real-world test data. 

\section{Related Work}

\noindent\textbf{GAN based data generation.}
GAN based data generation methods~\cite{hoffman2017cycada,Isola_2017_CVPR,zhang2018fully} focus on adjusting the style of synthetic images to approximate real-world images. 
For this kind of method, image (pixel)-level domain adaptation~\cite{Zhu_2017_ICCV} is a commonly used approach and has been proven to be effective in several works such as Pix2Pix \cite{Isola_2017_CVPR}, MUNIT \cite{huang2018munit}, WCT \cite{NIPS2017_6642} and SPGAN \cite{deng2018image}. 
To generate data that has similar appearance with target data, Zhang \emph{et al.}~\cite{zhang2018fully} use an appearance adaptation network to gradually generate a stylized image from a noise image by adapting the appearance to target data.
Chen \emph{et al.}~\cite{chen2019learning} further propose an input-level adaptation network that leverages the depth information to reconstruct the source image. It employs an adversarial learning \cite{NIPS2014_5423} framework to ensure style consistency between source and target domains.


\textbf{Graphic engine based data generation.}
Graphic engine based methods~\cite{Richter_2016_ECCV,gaidon2016virtual,Geiger2013IJRR} use the simulated 3D models, such as person~\cite{barbosa2018looking}, object~\cite{pepik2012teaching} and scene~\cite{satkin2012data,Geiger2013IJRR}, as well as varying virtual environments to render and generate synthetic data. 
On one hand, there is some works setting the conditions of generating data manually or randomly~\cite{Tremblay2018TrainingDN,prakash2019structured}. For example, Hattori \textit{et al.}~\cite{hattori2015learning} generate data by tuning the scene to match specific scene artificially to help detect pedestrians in real data. On the other hand, several recent researches propose to synthesize data by learning-based simulation methods~\cite{kar2019metasim,ruiz2018learning,Heusel2017GANsTB,prakash2019structured,Gretton2012}. For example, Ruiz \textit{et al.} \cite{ruiz2018learning} propose reinforcement learning-based method for adjusting the parameters of synthesized data to mimic KITTI~\cite{geiger2012we} and get significant performance improvement of synthetic data using learned parameters than random parameters. Yao \textit{et al.} propose to use FID \cite{Heusel2017GANsTB} metric with attribute descent to optimize the attributes of synthetic data for vehicle re-ID and get improved recognition results by using the optimized data than random attribute setting data.

\textbf{Learning from synthetic data.}
Based on the convenience of annotation of synthetic data, some datasets have been created to help learn models for related researches. 
For example, Richter \textit{et al.} \cite{Richter_2016_ECCV} create a pixel-level annotated dataset including 24,966 images through playing in the game Grand Theft Auto V for semantic segmentation.
Gaidon \textit{et al.} \cite{gaidon2016virtual} build a synthetic video dataset - Virtual KITTI that mimics KITTI \cite{Geiger2013IJRR} to support bigger benchmark datasets for the visual community.
Bak \textit{et al.}~\cite{bak2018domain} introduce a synthetic dataset SyRI including
100 characters with rich lighting conditions to learn robust person re-identification models for illumination variations. On the other hand, some work learns from synthetic data based on its controllability to investigate some problem of specific conditions~\cite{dosovitskiy2017carla,sun2019dissecting,sakaridis2018semantic}. For example, Sun \textit{et al.}~\cite{sun2019dissecting} discuss the influence of person viewpoint changes on person re-ID systems and Sakaridis \textit{et al.}~\cite{sakaridis2018semantic} build a Foggy Cityscapes dataset to learned semantic segmentation and object detection models that have improved performance on challenging real foggy scenes.

\section{SceneX: Complex Scene Generator}

Existing simulators \cite{dosovitskiy2017carla,prakash2019structured,tobin2017domain} are not very suitable to construct complex scenes for two reasons. First, the number of classes in the simulators is limited, so it is not feasible to perform segmentation on a rich range of real-world objects. Second, they define attribute at the instance level, \emph{e.g.,} the position and orientation of each object. Under this setting, we can only edit individual objects. Given a large number of objects, the editing space is huge and intractable.
In this section, we first describe the 3D assests in ScenceX corresponding to a standard 19 classes. We then introduce how SceneX is rendered in Unity, which allows us to optimize global attributes instead of instance-level attributes used in previous works.

\subsection{3D Scene Classes and Assets}
SceneX contains 19 classes of assets. Its classes are the same with Cityscapes \cite{cordts2016cityscapes}, \emph{i.e.,} car, pedestrian, building and \emph{etc}.
It is designed to generate street scenes from a first-person driver perspective. 
Specifically, 
SceneX contains 200 pedestrian models, 195 cars, 28 buses and 39 trucks from existing model repositories \cite{sun2019dissecting,yao2019simulating}, and makes necessary modifications so that they are compatible with our engine.
Besides, we collect 106 buildings, 18 bicycles, and 19 trees, among others. There are also 14 sky box models to simulate different weather conditions.
With these models, SceneX can generate complex scenes with a rich range of objects.

\begin{figure}
    \centering
    \includegraphics[scale=0.34]{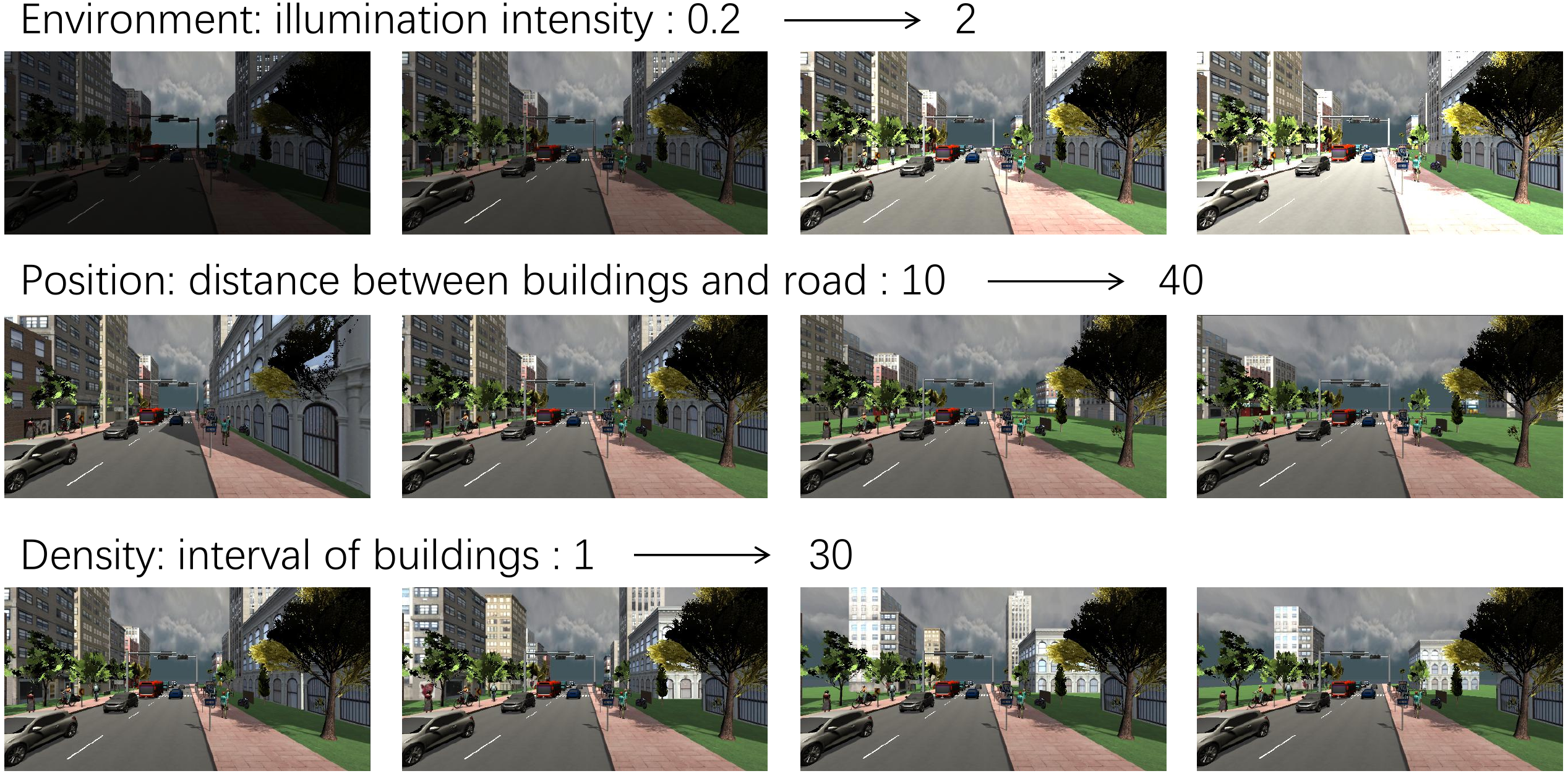}
    \caption{Example global attributes. (i) Illumination intensity changes the brightness of virtual environment, and (ii) distance between buildings and road changes the position of a group of buildings and (iii) interval of buildings changes the density of buildings.}
    \label{fig:attribute_list}
\end{figure}

\subsection{Engine Design and Global Attributes}\label{sec:global}
We aim to simulate scenes that contain many objects. Existing data simulation works generally are not very well designed to handle this problem. For example, Kar \emph{et al.} optimize the attributes of each object \cite{kar2019metasim}, which leads to a prohibitively large search space if the scene contains many objects. 
To accomplish our goal, we propose a different strategy regarding attribute manipulation. Details of engine design and global attributes are provided below. 

\textbf{Engine design.}
Our data synthesis engine is featured by a ``line-based'' design. To the center of a scene is a road map, around which 19 types of objects are placed. 
In the line-based design, the same types of objects (\emph{e.g.,} bicycles) are placed along a line parallel to the road, and thus these objects have the same distance to the road. 
Because objects on a line are tied, changing the position of an object means to move all the objects placed by the same magnitude on that line. 
This object placement strategy not only allows us to easily adjust the distance between objects and the road, but also enables precise object density changes by modifying the interval between objects on the same line. 
Therefore, when considering a single type of object (\emph{e.g.,} persons), its distribution within a scene is determined by its distance to the road and its density. 

\textbf{Global attributes.}
We use 23 global attributes to control the scene structure, including 8 for environment, 7 for object position and 8 for object density. We intuitively select these global attributes as they have large influence on the overall scene property. 
Examples of the global attributes are shown in Fig. \ref{fig:attribute_list}. 
Among them, \emph{illumination intensity} changes the brightness of virtual environment, which affects the visibility of objects. \emph{Distance between buildings and the road} changes the position of a group of buildings, and \emph{building interval} changes the density of buildings along the line. Fig. \ref{fig:attribute_list} shows that by editing the values of global attributes (only a few parameters), the scene structure / appearance can be significantly changed. 
The advantages of using global attributes is discussed in Section \ref{sec:discussion}.

As a controllable system, the attributes of SceneX are editable. 
Like \cite{yao2019simulating}, we build a Python API using Unity ML-Agents plugin \cite{DBLP:journals/corr/abs-1809-02627}. It allows us to modify the attributes directly through Python programming without needing expert knowledge about Unity.
We refer readers to the appendix for more details of our engine.


\section{Proposed Method}

\subsection{Problem Formulation}

Suppose we have a target dataset that is divided into two parts, \textit{i.e.}, a validation set $\bm{D}_v(\bm{X},\bm{Y})$ and a test set $\bm{D}_t(\bm{X},\bm{Y})$, where $\bm{X}$ and $\bm{Y}$ are a set of images and their segmentation labels, respectively.
Our objective is to train a policy network $G_{\bm\phi}$, parameterized by $\bm{\phi}$, which takes a set of randomly sampled attributes $\bm{\theta}$ as input and outputs a set of updated attributes $\bm{\theta}^{\prime}$.
Inputting $\bm\theta^\prime$ into SceneX, the engine will render a synthetic dataset $\bm{S}(\bm{X},\bm{Y})$, where $\bm{X}$ contains images, and $\bm{Y}$ represents the corresponding pixel-wise labels automatically acquired through rendering engine buffer. 
After training a segmentation network $T_{\bm\omega}$ (parameterized by $\bm\omega$) on $\bm{S}(\bm{X},\bm{Y})$ till convergence, we compute the accuracy on $\bm{D}_v(\bm{X},\bm{Y})$.
The accuracy score is used to update the policy network, enforcing it to increase the accuracy on $\bm{D}_v(\bm{X},\bm{Y})$, and thus on $\bm{D}_t(\bm{X},\bm{Y})$.
In short, the training process poses a bi-level optimization problem, \textit{i.e.},
\begin{subequations}
	\begin{equation}
	\bm{\phi}^* = \mathop{\arg\max}_{\bm{\phi}} \sum_{(\bm{x},\bm{y})\in \bm{D}_v(\bm{X},\bm{Y})} Score(\bm{y},T_{\bm{\omega}}(\bm{x};\bm{\omega}^*(\bm{\phi}))),
	\label{eqt:optimiz_psi}
	\end{equation}
	\begin{equation}
	s.t. \quad \bm{w}^*(\bm{\phi}) = \mathop{\arg\min}_{\bm{\omega}} \sum_{(\bm{x},\bm{y})\in \bm{S}(\bm{X},\bm{Y})} L(\bm{y}, T_{\bm{\omega}} (\bm{x};\bm{\omega}(\bm{\phi}))).
	\end{equation}
\end{subequations}
For two reasons, solving this problem with a gradient-based approach is not feasible. First, the mathematical rendering function of Unity is unknown. Second, the accuracy score is computed upon a training-till-convergence process. That is, we have to train the segmentation model till convergence before obtaining the segmentation accuracy as supervision signal. 

\subsection{Scalable Discretization-and-Relaxation}

In order to optimize the attributes of SceneX, we propose a scalable discretization-and-relaxation (SDR) optimization method. It follows a reinforcement learning framework, and employs a neural network to map random attribute values to updated ones.


\textbf{The reinforcement learning framework.}
Similar to \cite{kar2019metasim,ruiz2018learning}, our overall architecture adopts the REINFORCE algorithm \cite{williams1992simple} to tackle the system's being non-differentiable. 
The REINFORCE algorithm optimizes the problem through a sampling process, forcing it to maximize the following expected reward,
\begin{align}
	J(\bm{\phi}) = E_{\bm{\theta}^\prime \sim \pi_{\bm{\phi}}(\bm{\theta})}[Score(\bm{\theta}^\prime)],
\end{align}
with respect to $\bm\phi$. $Score(\bm\theta^\prime)$ is the accuracy score computed on the validation set $\bm{D}_v(\bm{X},\bm{Y})$, where $\bm\theta^\prime$ are updated attributes that are sampled from $G_{\bm\phi}(\bm\theta)$.
An unbiased, empirical estimate of the gradients for updating the policy is,
\begin{align}
	\nabla_{\bm\phi} J(\bm\phi) \approx \frac{1}{K}\sum_{k=1}^K \nabla_{\bm\phi} log(\pi_{\bm\phi}(\bm\theta))(Score(\bm\theta^\prime_k)-b),
\end{align}
where $b$ is a baseline that is usually chosen to be exponential moving average over previous rewards, and $K$ is the number of different datasets sampled in one policy.

\begin{figure}[t]
    \centering
    \includegraphics[scale=0.4]{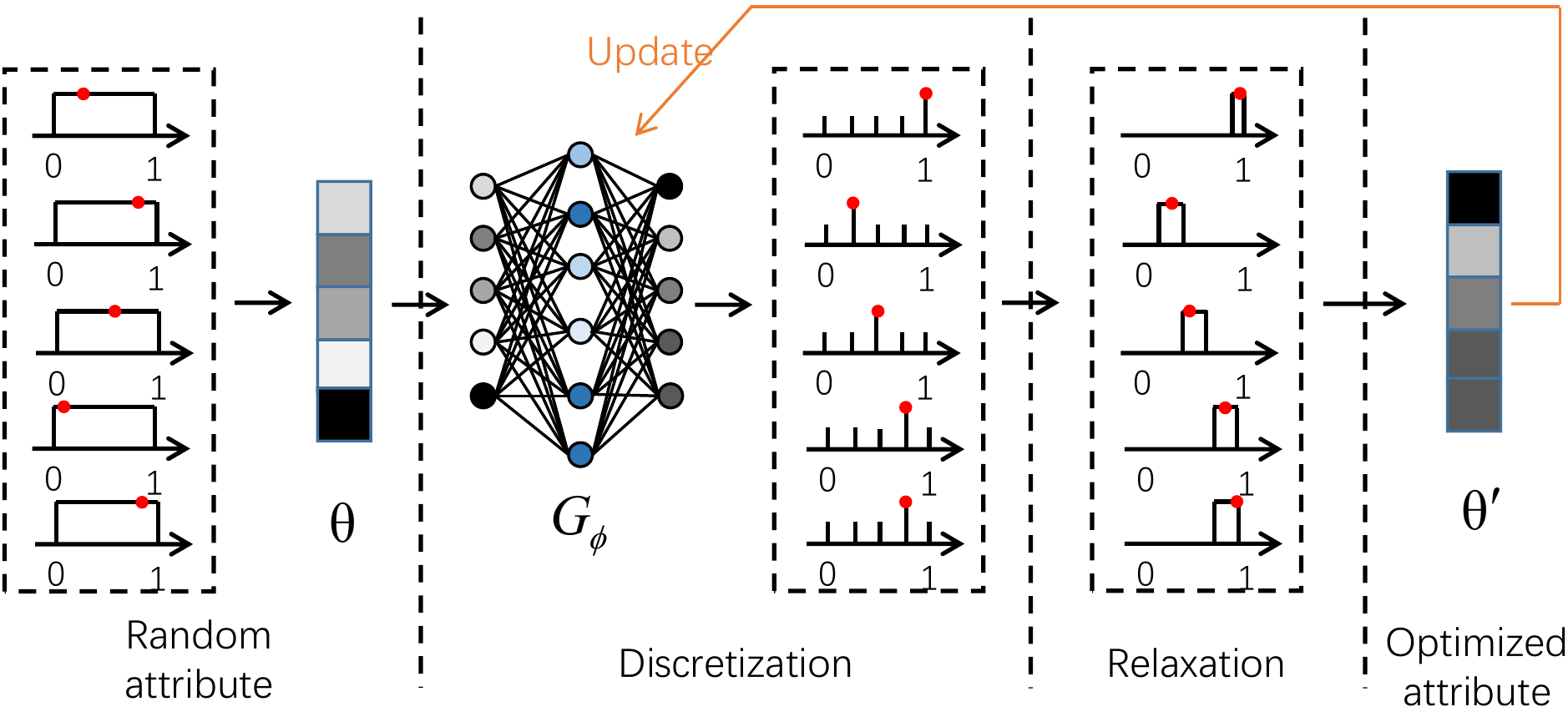}
    \caption{The proposed SDR method. Random attributes are fed into the policy network, outputting discrete values. Then we manually inject variance on top of these values. Final output values are used to update the policy network.}
    \label{fig:sdr}
\end{figure}

\textbf{Discretization-and-relaxation with MLP.}
We view this attribute optimization task as a distribution mapping problem that maps random attribute values to optimized ones.
In order to achieve our goal, we employ a multi-layer perceptron (MLP) to building a mapping function between random and updated attribute values.
The MLP optimizes the attribute values through a \textbf{discretization} process. Specifically, suppose the input of MLP is a $N$ dimensional vector that represents random attribute values, where $N$ denotes attribute number.
The corresponding output is a $N \times K$ dimensional vector with a softmax function applied to the second dimension, where $K$ denotes discrete number.
Under this form, we quantize each attribute into $K$ discrete values, with a probability distribution over $K$ numbers.
Each value is then sampled from the $K$ numbers during training, or determined by maximum probability during testing.
The sampled outputs are regarded as updated attribute values, with a dimension of $N$, and their probabilities are known so as to update the policy.
In order to remedy the loss of attribute diversity caused by discretization, a \textbf{relaxation} process is added afterwards, that is we manually inject variance on top of the updated attributes.
An overview of the proposed SDR method is illustrated in Fig. \ref{fig:sdr}.



\textbf{Method scalability.}
Our method is scalable in the following three aspects.
First, the attribute number $N$ is scalable, since we can change the input dimension. That means the change of attribute number to be optimized jointly each time.
Second, the discrete number $K$ is scalable, since we can change the output dimension. That means the change of refinement degree of output values for each attribute.
Third, the discrete number for each attribute is scalable, since we can assign different output layers for different attributes. That means to assign different discrete numbers (\emph{i.e.}, refinement degree) for different attributes.

\subsection{SDR in Groups}
As the number of attribute increases, the performance of REINFORCE algorithm drops rapidly due of the increasing sampling space.
In order to tackle it, we propose to use SDR in groups. That is, we split the attributes into several groups and optimize them using SDR in the form of coordinate descent \cite{wright2015coordinate}.

Specifically, the input of MLP is a $N$ dimensional vector $\bm{\bm\theta}$.
The corresponding output is a $N \times K$ dimensional vector $\bm\theta_{s}$, representing the sampling space of updated attributes.
In order to use SDR in groups, $\bm\theta$ is split into $n$ parts, \textit{i.e.}, $\bm\theta_{1}, \bm\theta_{2}, \dots, \bm\theta_{n}$, where $\bm\theta_{n}$ is a $N_n$ dimensional vector and $\sum^{i=0}_n N_i = N$.
The corresponding output vector is $\bm\theta_{s_1}, \bm\theta_{s_2}, \dots, \bm\theta_{s_n}$, where $\bm\theta_{s_n}$ is a $N_n \times K$ dimensional vector. 
Sampling from $\bm\theta_{s_n}$ can get $\bm\theta_{n}^\prime$ with dimension of $N_n$.
After concatenating $\bm\theta_{1}^\prime, \bm\theta_{2}^\prime, \dots, \bm\theta_{n}^\prime$, we can get the updated attributes $\bm\theta^\prime$.
The policy model $G_{\bm\phi}$ is also split into $n$ models during this process, resulting into $G_{\bm\phi_1}, G_{\bm\phi_2}, \dots G_{\bm\phi_n}$.
Each model is optimized under following equations:
\begin{equation}
    \begin{split}
        J_1 = &Score(concat[\bm\theta_{1}^\prime, \bm\theta_{2}, \dots, \bm\theta_{i}, \dots, \bm\theta_{n}]]), \\
        J_2 = &Score(concat[\bm\theta_{1}^\prime, \bm\theta_{2}^\prime, \dots, \bm\theta_{i}, \dots, \bm\theta_{n}]]),  \\
        & \qquad \qquad \dots \\
        J_i = &Score(concat[\bm\theta_{1}^\prime, \bm\theta_{2}^\prime, \dots, \bm\theta_{i}^\prime, \dots, \bm\theta_{n}]]),  \\
        & \qquad \qquad \dots \\
        J_n = &Score(concat[\bm\theta_{1}^\prime, \bm\theta_{2}^\prime, \dots, \bm\theta_{i}^\prime, \dots, \bm\theta_{n}^\prime]]).
    \end{split}
\end{equation}


\begin{figure}[t]
    \centering
    \includegraphics[scale=0.42]{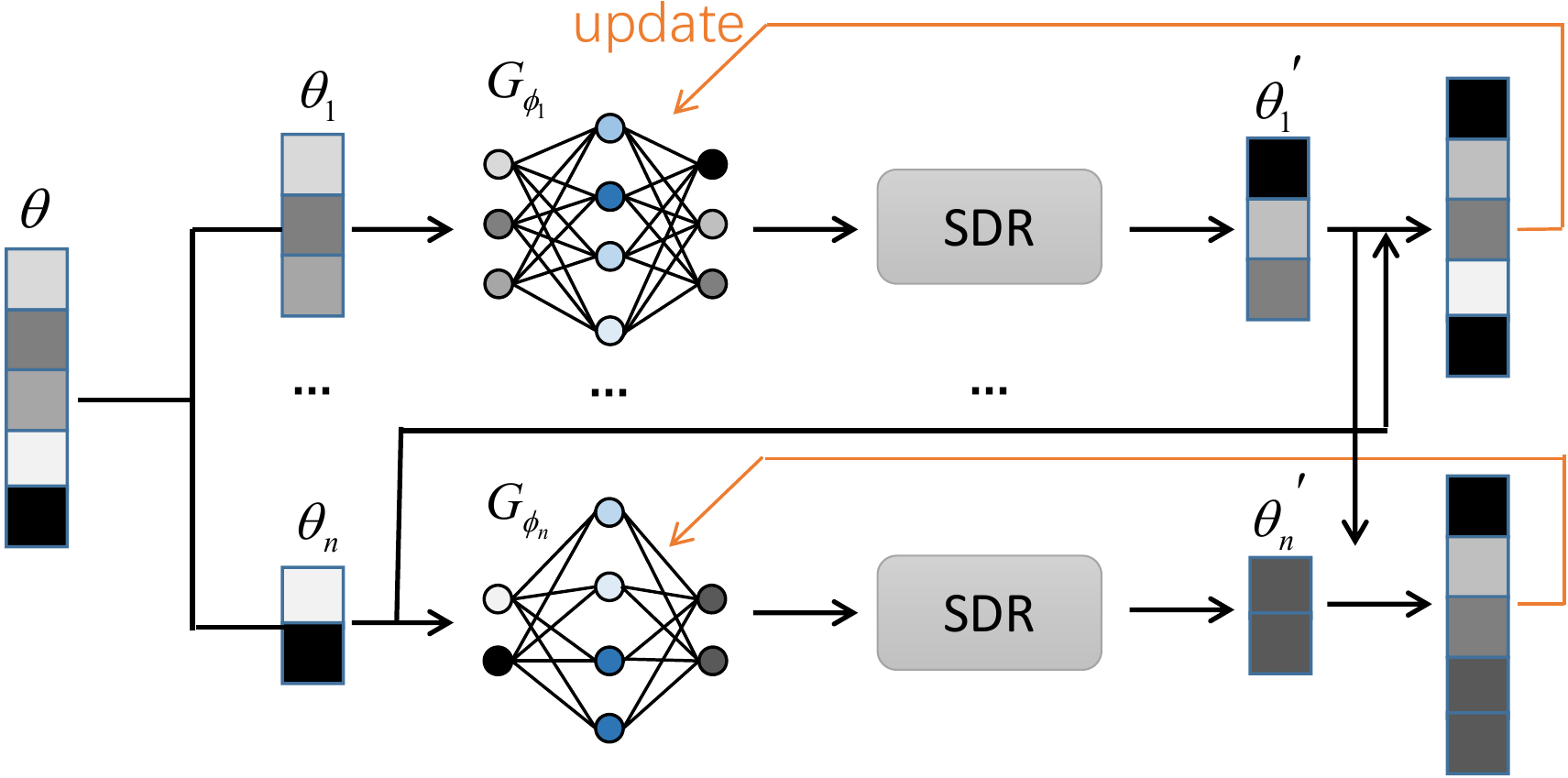}
    \caption{SDR in groups. We split the input, output and the policy network into several groups, and optimize each group by SDR using coordinate descent.}
    \label{fig:group}
\end{figure}

\subsection{Discussion}\label{sec:discussion}

\textbf{Difference with attribute descent~\cite{yao2019simulating}.} 
Attribute descent can be viewed as a special case of the proposed SDR. That is, if we replace the policy model in SDR with the brute force search and set the number of groups to $N$ ($N$ is the total number of attributes), our method will reduce to attribute decent. 
Attribute descent optimizes each attribute independently, while our method jointly optimizes attributes in a group, thus considering their dependencies. Besides, the algorithm complexity of attribute descent is $O(NK)$, so it is time-consuming when the number of attributes is large. In comparison, the complexity of SDR is $O(nK)$, where $n$ is the group number, which is much smaller. 


\textbf{Difference with learning to simulate~\cite{ruiz2018learning} and meta-sim~\cite{kar2019metasim}.} 
The difference with these two methods lies in the usage of policy. In \cite{ruiz2018learning,kar2019metasim}, A Gaussian model is used as the policy, where the parameters to be learned are the mean and variance.
This strategy requires the policy to sample within a large range value, so it is very sensitive to the initial value.
Moreover, using the Gaussian model means that only continuous attributes can be optimized.
As a result, these two methods perform sampling in a huge search space.
Moreover, our method departs from these two methods in that we use global attributes rather than instance-level ones. The advantages of using global attributes are discussed below. 

\textbf{Why global attributes?} 
The advantages of using global attributes over instance-level attributes are two-fold. First, 
global attributes more directly represent the characteristics of a scene. For example, by directly manipulating the density of pedestrians and cars, urban and rural areas can be better characterized. By decreasing the distance between buildings and road, we can directly mimic the situation in a modern city. In comparison, manipulating the location of individual cars and persons apparently gives much less direct impact on the overall structure and significantly increases system computational burden. 

Second, the search space of global attribute is much smaller.
Suppose all the objects are placed on a two-dimensional map. The search space for an individual object is $S_x \cdot S_y$, where $S_x$ and $S_y$ represent the search space (in pixels) along $x$ and $y$ axis, respectively. Suppose a scene has $C$ classes of objects, and that $N_i$ is number of objects for each class. The search space of the scene is $\prod_{i=1}^{C} (S_x \cdot S_y)^{N_i}$ if the scene structure optimization is performed under instance level. In comparison, for global-level attribute optimization, the search space is significantly reduced to $({S_x \cdot D})^C$ since we place the same type of objects on a line, where $D$ represents the range of object density and is smaller than $S_y$. A reduced search space allows our method to operate efficiently and converge to a superior state.

\section{Experiment}

In this section, we first compare SDR with other attribute optimization methods. Then, we compare the effectiveness as a training set of our simulated images with existing synthetic datasets GTA5 and SYNTHIA. Besides, we show that our simulated data is beneficial for pre-training. Finally, we verify the necessity of each component in the proposed SDR method.

\begin{table*}[t]
    \centering
    \caption{Segmentation accuracy on (a) Cityscape and (b) CamVid datasets. Four synthetic datasets are used for training, \emph{i.e.,} SYNTHIA, GTA5, SceneX by RA (random attributes), and SceneX by SDR (ours). Two networks are used, FCN8s and DeepLabv2. We highlight the numbers where our method SDR gives the highest accuracy for the corresponding class.}
    \begin{tabular}{l|l|p{0.5cm}<{\centering}|p{0.3cm}<{\centering}p{0.3cm}<{\centering}p{0.3cm}<{\centering}p{0.3cm}<{\centering}p{0.3cm}<{\centering}p{0.3cm}<{\centering}p{0.3cm}<{\centering}p{0.3cm}<{\centering}p{0.3cm}<{\centering}p{0.3cm}<{\centering}p{0.3cm}<{\centering}p{0.3cm}<{\centering}p{0.3cm}<{\centering}p{0.3cm}<{\centering}p{0.3cm}<{\centering}p{0.3cm}<{\centering}p{0.3cm}<{\centering}p{0.3cm}<{\centering}p{0.35cm}<{\centering}|c}
\multicolumn{22}{c}{(a) Synthetic dataset $\rightarrow$ Cityscapes} \\
\hline
Dataset & Size  & Net & \rotatebox{90}{Road}  & \rotatebox{90}{SW}  & \rotatebox{90}{Build}  & \rotatebox{90}{Wall}   & \rotatebox{90}{Fence}   & \rotatebox{90}{Pole}   & \rotatebox{90}{TL}  &\rotatebox{90}{TS}  &\rotatebox{90}{Veg}  &\rotatebox{90}{Terr}   &\rotatebox{90}{Sky}  &\rotatebox{90}{PR}  &\rotatebox{90}{Rider}  &\rotatebox{90}{Car}  &\rotatebox{90}{Truck}  &\rotatebox{90}{Bus}  &\rotatebox{90}{Train}  &\rotatebox{90}{Motor}  &\rotatebox{90}{Bike}   & mIoU \\
\hline
\hline
SYNTHIA & 9.4k & \multirow{4}{*}{\rotatebox{90}{FCN8s}} &37.0 &22.8 &63.5 &0.1 &0 &4.8 &0 &0 &71.1 &0 &73.1 &35.1 &4.6 &25.7 &0 &6.4 &0 &0 &0 &18.1 \\
GTA5 & 24.9k & &34.3 &16.3 &69.2 &12.8 &12.0 &7.7 &0 &0 &75.7 &15.7 &65.5 &26.9 &0 &38.2 &10.4 &1.8 &0 &0 &0 &20.3 \\
SceneX+RA & 450 & &51.6 &17.8 &48.4 &0 &0 &1.5 &0.8 &0 &61.9 &6.2 &13.5 &0.3 &1.5 &0.1 &1.3 &2.1 &0 &0 &2.0 &10.9 \\
SceneX+SDR  & 450 &  & \textcolor{blue}{66.1} &\textcolor{blue}{25.0} &56.3 &0.5 &0 &\textcolor{blue}{8.9} &0.5 &0 &68.8 &7.0 &41.7 &21.5 &\textcolor{blue}{10.6} &28.8 &6.0 &0.3 &0 &0 &\textcolor{blue}{12.6} &18.6 \\
\hline
SYNTHIA &9.4k & \multirow{4}{*}{\rotatebox{90}{DeepLabv2}} &45.4 &18.6 &66.8 &15.2 &10.8 &16.6 &11.6 &0.6 &77.1 &16.7 &65.3 &39.6 &2.1 &49.9 &8.9 &11.6 &0 &5.5 &0 &18.3 \\
GTA5 & 24.9k  & &12.3 &19.9 &40.8 &1.6 &0 &15.5 &0.7 &3.9 &75.3 &0 &70.6 &38.3 &2.4 &44.9 &0 &14.0 &0 &0.2 &6.6 &24.2 \\
SceneX+RA & 450 & &63.8 &10.4 &56.1 &0.1 &0 &3.1 &2.1 &0.1 &57.1 &0.9 &20.7 &5.2 &1.7 &6.4 &1.4 &1.6 &0.3 &0.9 &1.1 &12.3 \\
SceneX+SDR & 450 &  & \textcolor{blue}{70.6} &18.6 &63.1 &4.0 &0.4 &10.4 &0.2 &0.7 &64.2 &3.6 &40.6 &27.1 &\textcolor{blue}{3.7} &30.0 &2.0 &0 &0 &1.2 &\textcolor{blue}{8.5} &18.4 \\ 
\hline
\end{tabular}

\vspace{2pt}
\begin{tabular}{l|l|p{0.5cm}<{\centering}|p{0.7cm}<{\centering}p{0.7cm}<{\centering}p{0.7cm}<{\centering}p{0.7cm}<{\centering}p{0.7cm}<{\centering}p{0.7cm}<{\centering}p{0.7cm}<{\centering}p{0.7cm}<{\centering}p{0.7cm}<{\centering}p{0.7cm}<{\centering}p{0.7cm}<{\centering}|c}
\multicolumn{14}{c}{ (b) Synthetic dataset $\rightarrow$ CamVid} \\
\hline
Dataset & Size & Net & Sky  & Build  & Pole  & Road   & SW   & Tree   & Sign  &Fence  &Car  &PR   &Bike  & mIoU \\
\hline
\hline
SYNTHIA & 9.4k & \multirow{4}{*}{\rotatebox{90}{FCN8s}} &81.6 &65.2 &0.7 &63.6 &42.1 &47.4 &0 &0 &46.9 &19.1 &0 &33.3 \\
GTA5 & 24.9k & &75.6 &67.8 &0 &66.4 &43.3 &56.8 &0 &0 &53.8 &0 &0 &33.1 \\
SceneX+RA & 450 & &42.7 &56.0 &1.1 &35.9 &44.1 &41.3 &0 &0 &25.3 &0 &0 &22.4 \\
SceneX+SDR & 450 &  & \textcolor{blue}{81.7} &53.7 &\textcolor{blue}{2.6} &\textcolor{blue}{69.3} &\textcolor{blue}{44.9} &29.4 &\textcolor{blue}{1.0} &0 &24.7 &10.7  &\textcolor{blue}{5.7} &29.6 \\
\hline
SYNTHIA & 9.4k & \multirow{4}{*}{\rotatebox{90}{DeepLabv2}} &65.6 &62.2 &10.3 &55.3 &36.1 &47.6 &1.6 &0 &50.1 &27.6 &4.6 &32.6 \\
GTA5 & 24.9k  & &58.3 &63.4 &7.5 &34.5 &31.3 &53.8 &11.7 &22.2 &65.9 &10.1 &0 &32.6 \\
SceneX+RA & 450 & & 64.6 &63.5 &2.3 &55.1 &36.4 &44.1 &4.5 &0 &33.6 &0 &2.1 &27.8 \\
SceneX+SDR & 450 &  & \textcolor{blue}{68.1} &55.6 &7.2 &\textcolor{blue}{60.1} &\textcolor{blue}{45.8} &42.1 &\textcolor{blue}{12.2} &0 &35.7 &14.2 &4.2 &31.4 \\
\hline
\end{tabular}
    \label{tab:dataset_comp}
\end{table*}

\subsection{Experimental Setting}

\textbf{Datasets for attribute training and testing.}
We use two real-world datasets to train the attributes of SceneX.
The \textbf{Cityscapes} dataset \cite{cordts2016cityscapes} contains 2,975 images in the training set and 500 images in the validation set, all of size $2,048 \times 1,024$. We select 500 images from the training set for attribute training and calculate model accuracy on the validation set.
We down-sample the images to $640 \times 320$ during attribute training and $1,024 \times 512$ during testing.
For the pre-training experiment in Section \ref{sec:pretrain}, we fine-tune the pre-trained network on the training set at image resolution of $1,024 \times 512$, and report the results on the validation set at original size.
The \textbf{CamVid} dataset \cite{brostow2008segmentation} contains 367 and 233 images for training and testing, respectively. We use the training set for attribute training and compute model accuracy on the test set. The dataset images have a fixed spatial resolution of $960 \times 720$, and we down-sample them to $480 \times 360$ at all settings.

\textbf{Datasets for Comparison.}
We compare our simulated dataset (named UnityScene) with two existing synthetic datasets, GTA5 \cite{Richter_2016_ECCV} and SYNTHIA \cite{ros2016synthia}.
GTA5 consists of 24,966 images with resolution of $1,914\times1,052$ obtained from the GTA5 video game. 
The ground truth annotations are compatible with the Cityscapes dataset that contains 19 categories. SYNTHIA \cite{ros2016synthia} is a dataset with synthetic images of urban scenes. 
The rendering covers a variety of environments and weather conditions. We adopt the SYNTHIA-RAND-CITYSCAPES subset that contains 9,400 images. The 19 categories in UnityScene, GTA5 and SYNTHIA are consistent.

\textbf{Evaluation metric.}
We use the commonly used mean intersection over union (mIoU) as evaluation metric.

\textbf{Implementation details.}
For the policy network, we deploy a three-layer MLP with hidden dimension of 256, and output dimension ($K$) of 10.
We use the Adam optimizer with a fixed learning rate of $1\times10^{-2}$.
The 23 attributes are first permutated and then manually grouped into 2-8 attributes in each group. 
Details of attribute grouping can be accessed in the supplementary material.

For the segmentation model, we deploy the widely used FCN8s \cite{long2015fully} and DeepLabv2 \cite{chen2017deeplab}, which both adopt VGG16 \cite{simonyan2014very} as backbone. During training, we use the SGD optimizer with a base learning rate of $5\times 10^{-4}$. Following Zhao \textit{et al.} \cite{Zhao_2017_CVPR}, we deploy the poly learning rate decay by multiplying the factor $(1-\frac{iter}{total-iter})^{0.9}$. 

We compute the accuracy score on the real-world validation set after training the segmentation model for 1000 iterations on simulated images and then update the policy network, which repeats for 50 times.
It takes 2.2 and 3.1 seconds to obtain an image and its segmentation label for a spatial resolution of $640 \times 320$ and $1024 \times 512$ respectively on an AMD Ryzen Threadripper 2950X CPU.
Besides rendering, we use one RTX 2080Ti GPU for deep learning experiment.
The simulated dataset size is 180 in training process.

\begin{figure}[t]
    \centering
    \includegraphics[scale=0.5]{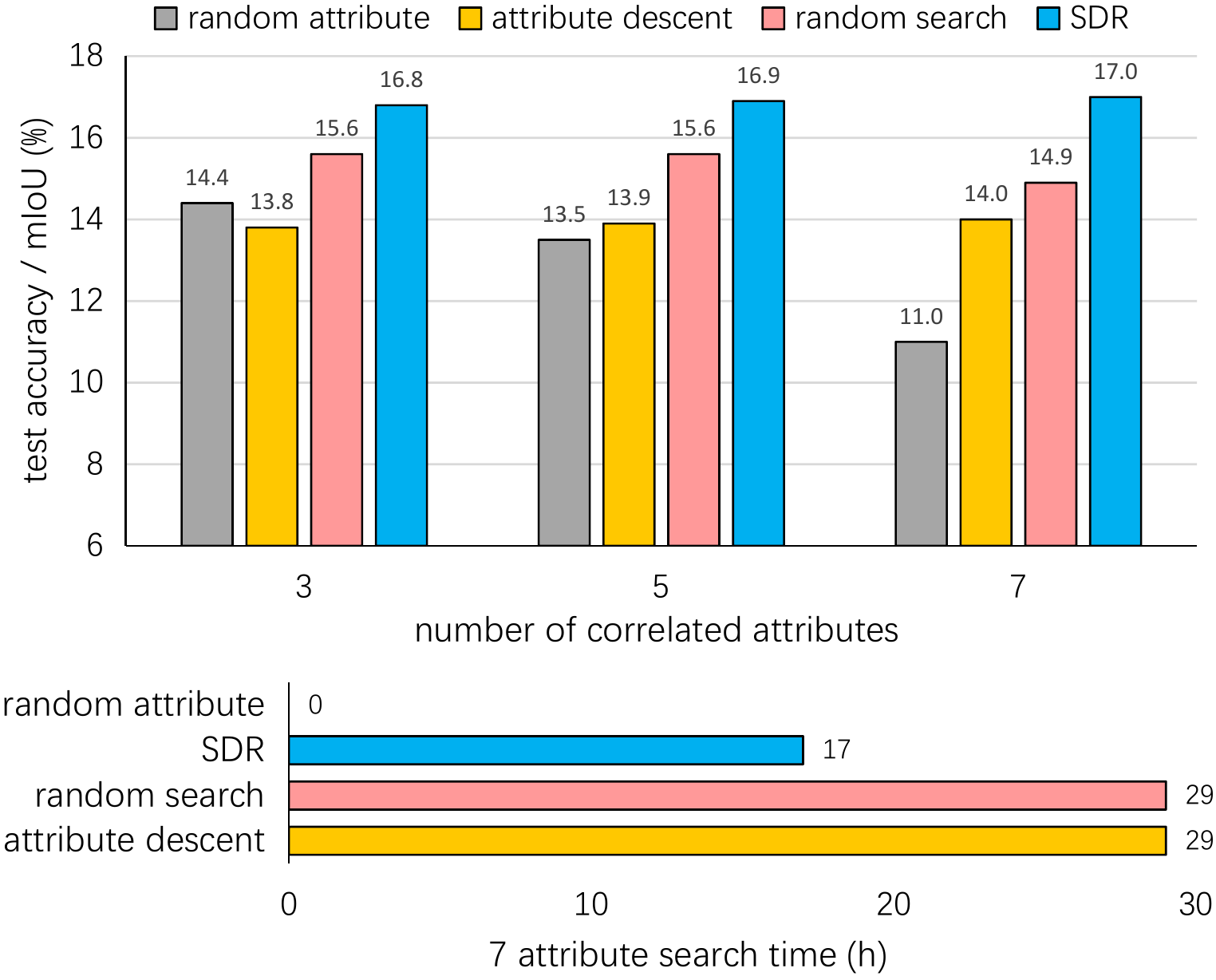}
    \caption{Method comparison on the Cityscapes validation set. (\textbf{Top:}) we choose to optimize 3, 5, and 7 correlated attributes, respectively. (\textbf{Bottom:}) we report the search time of optimizing 7 attributes. Four attribute learning methods are compared. SDR gives the best accuracy while consuming much less time than attribute descent and random search.}
    \label{fig:method_compare}
\end{figure}

\subsection{Comparative Study}\label{sec:comparative_study}

\textbf{Effectiveness of SDR over the random attributes baseline.} We render datasets using random attributes and attributes optimized by SDR, respectively. We train FCN8s and DeepLabv2 on these datasets and compare the model performance on Cityscapes and CamVid datasets. Results are summarized in Table \ref{tab:dataset_comp} and Fig. \ref{fig:method_compare}.

\begin{figure*}[t]
    \centering
    \includegraphics[scale=0.6]{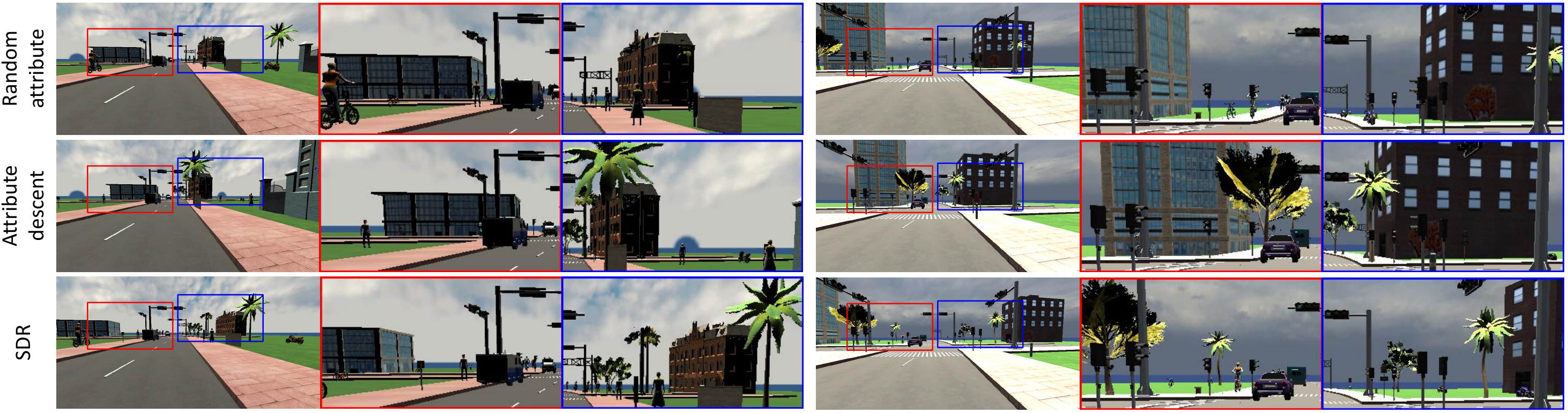}
    \caption{Examples of generated synthetic images by random attribute, attribute descent and SDR within SceneX. Random attribute randomizes object positions within a large range. Attribute descent tends to places visually obvious objects (\emph{e.g.,} building, tree) close to the road, resulting in severe overlap. In comparison, we observe that SDR place objects at more reasonable positions, such as person, rider on the sidewalk, tree on the terrain, and building away from the road for less occlusion.}
    \label{fig:generated_images}
\end{figure*}

It is clear from Table \ref{tab:dataset_comp} and Fig. \ref{fig:method_compare} that attributes learned through SDR are significantly superior to random attributes in synthesizing effective datasets. For example, on the Cityscapes dataset, the mIoU produced by our method (SceneX+SDR) is higher than random attributes (SceneX+RA) by +7.7\% and +6.1\% using FCN8s and DeepLabv2, respectively. Such an advantage exists in most classes. A similar trend can be observed on CamVid. 

\textbf{Comparing SDR with other attribute optimization methods.} 
In Fig. \ref{fig:method_compare}, we compare SDR with several attribute learning methods, including attribute descent \cite{yao2019simulating} and random search.
Random search samples many set of attributes and gets the best attribute combination by brute force search.
Because the compared methods are considered not scalable \emph{w.r.t} the number of attributes, this experiment will optimize a fraction of the total 23 attributes. Specifically, we select 7 correlated attributes, \emph{i.e.}, 7 object position attributes (see Supplementary Material for details). Among them, we select 3, 5, and 7 attributes, forming three sets of experiment. 


From the perspective of \textbf{segmentation accuracy}, we observe that SDR consistently outperforms the competing algorithms. Attribute descent does not consider attribute correlations and gives the lowest accuracy among the compared methods. It is even on par with random attributes baseline when optimizing 3 attributes. It indicates that when synthesizing complex scenes, it is of vital importance to consider attribute correlations, because various types of objects are closely related in the scene structure. In this regard, both SDR and random search consider attribute correlation via joint optimization.
The difference is that random search faces a large search space (it does not have the grouping operation). Therefore, as the number of attributes increases, it gets harder for random search to find an appropriate attribute combination, so the performance gap between SDR and random search is larger under 7 attributes compared with 3 and 5 attributes. 

From the view of \textbf{efficiency}, our optimization method converges faster than attribute descent and random search (saving 40\% time), due to the discretization and grouping operations. Specifically, when optimizing 7 attributes, the time needed for SDR, attribute descent, and random search is 17h, 29h, 29h, respectively. When optimizing 23 attributes, our method takes 52h while time for the other two methods will increase proportionally. 


We show examples of synthetic images using different methods within SceneX in Fig. \ref{fig:generated_images}. Random attribute randomizes object positions within a large range. Attribute descent tends to places visually obvious objects (\emph{e.g.,} buildings, trees) close to the road, resulting in severe overlap and a crowded scene. In comparison, SDR finds a more appropriate attribute combination, resulting in a more reasonable scene. For example, SDR places pedestrians and riders on the sidewalk, and trees on the terrain.

\textbf{Comparing optimized SceneX with GTA5 and SYNTHIA as effective training sets.} We respectively use GTA5, SYNTHIA, and SceneX (optimized by SDR) as training data, and use Cityscapes and CamVid as testing data in Table \ref{tab:dataset_comp}. 
We observe that SceneX (by SDR) produces promising accuracy: very competitive on Cityscapes compared with SYNTHIA, and slightly lower than SYNTHIA and GTA5 on CamVid. 
In important classes such as \emph{road}, \emph{bicycle} and \emph{rider}, SceneX exhibits the highest segmentation accuracy.

For two understandable reasons, models trained with SceneX are not superior to those trained with SYNTHIA and GTA5. First, as shown in Table \ref{tab:dataset_comp}, SceneX only contains 450 images, much fewer than 24,966 and 9,400 in GTA5 and SYNTHIA, respectively. This is because SceneX has a limited number of 3D assests, which cannot support the content diversity of a database as large as several thousand. Second, because GTA5 and SYNTHIA images are collected from video games that were carefully designed by professionals, their 3D assests are much more realistic than SceneX. These two limitations will be addressed in our next version by including more diverse and realistic 3D models. Here we emphasize the advantages of SceneX is unparalleled: SYNTHIA and GTA5 only contain static images, while SceneX content can be freely edited. The strength of such content editability is obvious: only 450 images can provide very promising segmentation accuracy on real-world datasets. 


\begin{figure}[t]
    \centering
    \includegraphics[scale=0.55]{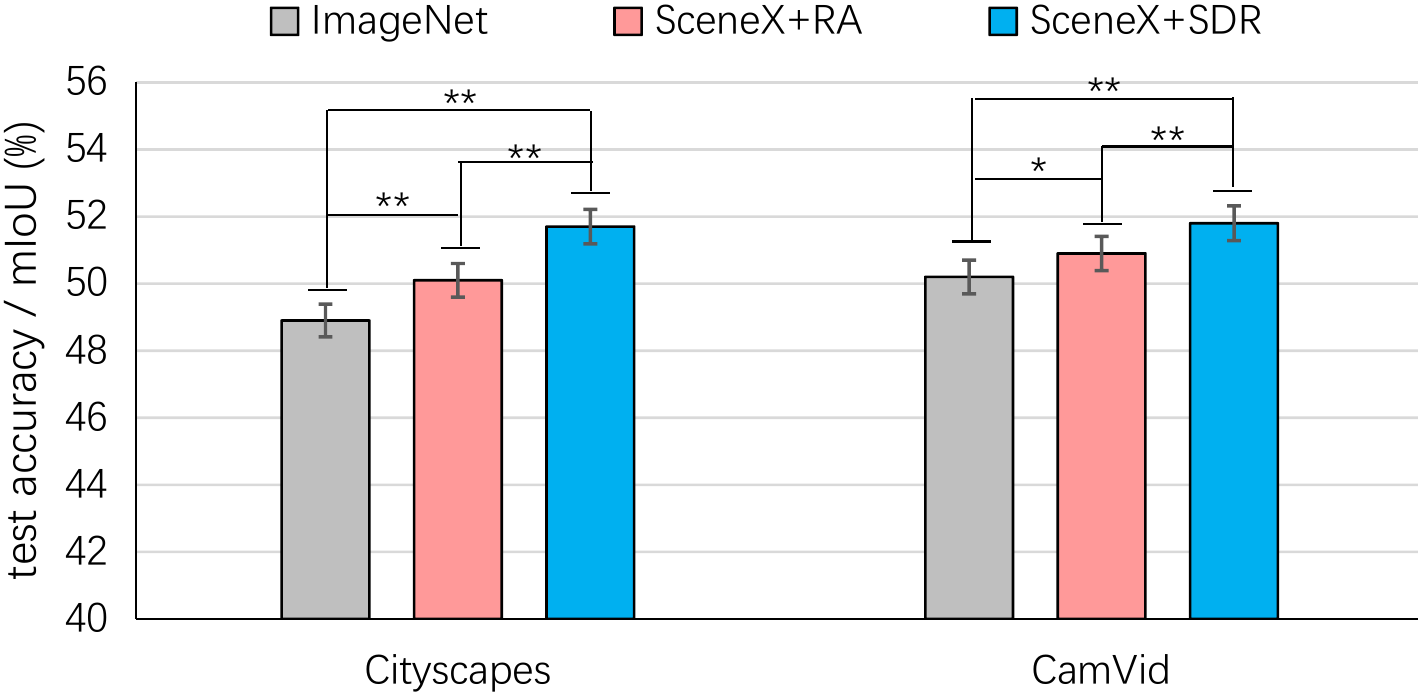}
    \caption{Evalutation of dataset abilities in pre-training. We compare ImageNet, SceneX+RA, and SceneX+SDR. The model is pretrained on synthetic data and fine-tuned and tested on Cityscapes and CamVid. We use statistical significance analysis to show the training stability. ``*'' means statistically significant (\emph{i.e.,} 0.01<p-value<0.05) and ``**'' means statistically very significant (\emph{i.e.,} p-value<0.01) respectively.}
    \label{fig:pretrain}
\end{figure}

\subsection{Simulation as Pre-training}\label{sec:pretrain}
Here we compare SceneX (SDR) with ImageNet and SceneX (random attributes, RA) their ability in model pretraining. We use the FCN8s as the segmentation model. Model fine-tuning is performed on the Cityscapes and CamVid datasets, respectively. Results are shown in Fig. \ref{fig:pretrain}.
The results indicate that using SceneX+SDR for pre-training yields higher accuracy than ImageNet as well as SceneX+RA. This comparison is statistically significant. 
Besides, SceneX with random attributes also yields statistically higher accuracy than ImageNet.
These results suggest that synthetic data optimized towards the target domain (\emph{i.e.,} Cityscapes or CamVid) have the potential to be a more effective source for model pre-training. 

\begin{figure}[t]
    \centering
    \includegraphics[scale=0.55]{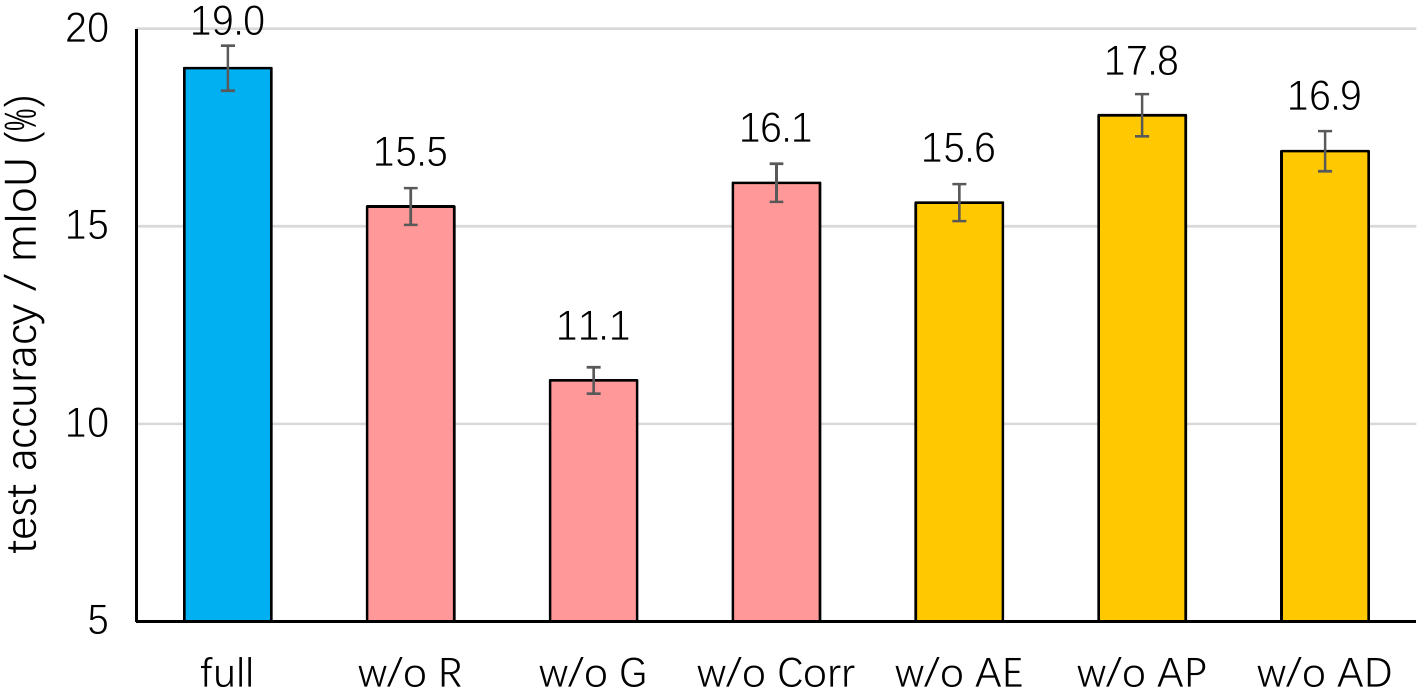}
    \caption{Ablation study for the SDR method. We remove the following components one at a time: relaxation (R), attribute grouping (G), considering attribute correlation (Corr), attributes for environments (AE), attributes for object position (AP), and attributes for object density (AD).}
    \label{fig:ablation_study}
\end{figure}

\subsection{Ablation Study}
In this section, we present the necessity of individual component in SDR.
The details of ablation study are shown in Fig.\ref{fig:ablation_study}.

\textbf{Relaxation is necessary.} It adds variance to the discrete attribute values and thus increases the diversity of generated scenes. Removing the relaxation process leads to an mIoU drop of 3.5\%.

\textbf{Optimizing attributes in groups is beneficial.} 
Without grouping, the search space becomes very large, and the algorithm may fall into inferior local optimums, causing the mIoU to drop by 7.9\%.

\textbf{Necessity of considering attribute correlation.} Without manually grouping correlated attributes into the same group, mIoU will drop from 19.0\% to 16.1\%. Fig. \ref{fig:generated_images} shows that unreasonable scene structures will be generated in this case. 

\textbf{Importance of different types of global attributes.} 
Three types of attributes are optimized: those related to environment, object location and object density. If we remove each attribute category (they have 8, 7, and 8 attributes, respectively), the mIoU will drop by 3.4\%, 1.2\% and 2.1\%, respectively. It indicates that the imaging condition, scene layout (object position) and density are essential to determining scene content.



\begin{figure}
    \centering
    \includegraphics[scale=0.6]{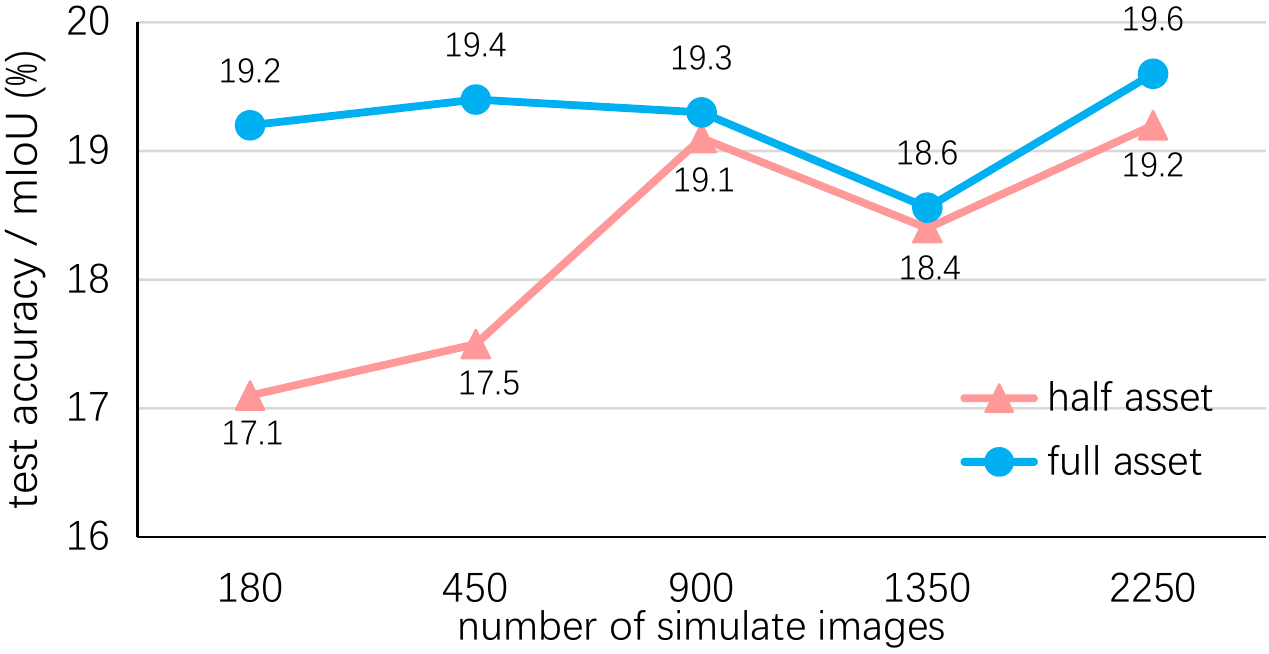}
    \caption{Impact of the number of 3D assets and the number of simulated images on test accuracy. We report mIoU (\%) on the Cityscapes validatoin set.}
    \label{fig:asset_and_img}
\end{figure}


\subsection{Important Parameters}
Here, we analyze the impact of some important parameters in our data simulation method. 
The parameters include the number of 3D assets and simulated images.

As shown in Fig. \ref{fig:asset_and_img}, using all of the 3D assets within SceneX (full asset) always obtains higher accuracy than using half of the assets (half asset). This indicates the importance of the number of assets used in our engine. Since SceneX is extendable, it is beneficial to improve segmentation accuracy by adding more 3D models.

Besides, the number of simulated images also matters. Firstly, simulating few images (\emph{e.g.,} 180, 450) in case of half asset is harmful to the test accuracy. This may be caused by the lack of some good 3D models in this case. Secondly, as the number of simulates images increases, the test accuracy tends to be slightly increasing, except the case of 1350 images, which we guess is influenced by the training process. Overall, the test accuracy is stable around 19 \% mIoU, which is quite a promising real-world segmentation accuracy compared with existing synthetic datasets.

\section{Conclusions}

Due to its convenience of acquiring ground truth labels at a large scale, data simulation is becoming a promising solution to the problem of lacking annotated data. 
Simulated data offers us a unique opportunity in content adaptation, \emph{i.e.,} editing image content to generate a training set useful for the target domain.
This paper proposes a scalable solution towards complex scene synthesis to be utilized in training semantic segmentation models. 
Our contribution is two-fold. 
First, we introduce a new 3D scene generation engine - SceneX, which construct scenes based on global-level attributes, such as illumination and object density.
Second, our solution explicitly considers attribute correlation, and its structure follows a discretization-and-relaxation strategy, making it uniquely suitable for the challenging scene generation problem at hand. 
We show that our optimized dataset is consistently superior to that generated by random attributes. With only 450 images, the optimized SceneX dataset is very close to the performance of GTA5 and SYNTHIA that have many thousands of realistic images. 
These results strongly support the idea of content adaptation. 
In future, we will collect more diverse and more realistic 3D models and dive deeper into this interesting area. 



\bibliographystyle{ACM-Reference-Format}
\bibliography{refs}


\begin{thebibliography}{50}


\ifx \showCODEN    \undefined \def \showCODEN     #1{\unskip}     \fi
\ifx \showDOI      \undefined \def \showDOI       #1{#1}\fi
\ifx \showISBNx    \undefined \def \showISBNx     #1{\unskip}     \fi
\ifx \showISBNxiii \undefined \def \showISBNxiii  #1{\unskip}     \fi
\ifx \showISSN     \undefined \def \showISSN      #1{\unskip}     \fi
\ifx \showLCCN     \undefined \def \showLCCN      #1{\unskip}     \fi
\ifx \shownote     \undefined \def \shownote      #1{#1}          \fi
\ifx \showarticletitle \undefined \def \showarticletitle #1{#1}   \fi
\ifx \showURL      \undefined \def \showURL       {\relax}        \fi
\providecommand\bibfield[2]{#2}
\providecommand\bibinfo[2]{#2}
\providecommand\natexlab[1]{#1}
\providecommand\showeprint[2][]{arXiv:#2}

\bibitem[\protect\citeauthoryear{Bak, Carr, and Lalonde}{Bak
  et~al\mbox{.}}{2018}]%
        {bak2018domain}
\bibfield{author}{\bibinfo{person}{Slawomir Bak}, \bibinfo{person}{Peter Carr},
  {and} \bibinfo{person}{Jean-Francois Lalonde}.}
  \bibinfo{year}{2018}\natexlab{}.
\newblock \showarticletitle{Domain adaptation through synthesis for
  unsupervised person re-identification}. In
  \bibinfo{booktitle}{\emph{Proceedings of the European Conference on Computer
  Vision}}.
\newblock


\bibitem[\protect\citeauthoryear{Barbosa, Cristani, Caputo, Rognhaugen, and
  Theoharis}{Barbosa et~al\mbox{.}}{2018}]%
        {barbosa2018looking}
\bibfield{author}{\bibinfo{person}{Igor~Barros Barbosa}, \bibinfo{person}{Marco
  Cristani}, \bibinfo{person}{Barbara Caputo}, \bibinfo{person}{Aleksander
  Rognhaugen}, {and} \bibinfo{person}{Theoharis Theoharis}.}
  \bibinfo{year}{2018}\natexlab{}.
\newblock \showarticletitle{Looking beyond appearances: Synthetic training data
  for deep cnns in re-identification}.
\newblock \bibinfo{journal}{\emph{Computer Vision and Image Understanding}}
  \bibinfo{volume}{167} (\bibinfo{year}{2018}), \bibinfo{pages}{50--62}.
\newblock


\bibitem[\protect\citeauthoryear{Brostow, Fauqueur, and Cipolla}{Brostow
  et~al\mbox{.}}{2009}]%
        {brostow2009semantic}
\bibfield{author}{\bibinfo{person}{Gabriel~J Brostow}, \bibinfo{person}{Julien
  Fauqueur}, {and} \bibinfo{person}{Roberto Cipolla}.}
  \bibinfo{year}{2009}\natexlab{}.
\newblock \showarticletitle{Semantic object classes in video: A high-definition
  ground truth database}.
\newblock \bibinfo{journal}{\emph{Pattern Recognition Letters}}
  \bibinfo{volume}{30}, \bibinfo{number}{2} (\bibinfo{year}{2009}),
  \bibinfo{pages}{88--97}.
\newblock


\bibitem[\protect\citeauthoryear{Brostow, Shotton, Fauqueur, and
  Cipolla}{Brostow et~al\mbox{.}}{2008}]%
        {brostow2008segmentation}
\bibfield{author}{\bibinfo{person}{Gabriel~J Brostow}, \bibinfo{person}{Jamie
  Shotton}, \bibinfo{person}{Julien Fauqueur}, {and} \bibinfo{person}{Roberto
  Cipolla}.} \bibinfo{year}{2008}\natexlab{}.
\newblock \showarticletitle{Segmentation and recognition using structure from
  motion point clouds}. In \bibinfo{booktitle}{\emph{European conference on
  computer vision}}. Springer, \bibinfo{pages}{44--57}.
\newblock


\bibitem[\protect\citeauthoryear{Chen, Papandreou, Kokkinos, Murphy, and
  Yuille}{Chen et~al\mbox{.}}{2017}]%
        {chen2017deeplab}
\bibfield{author}{\bibinfo{person}{Liang-Chieh Chen}, \bibinfo{person}{George
  Papandreou}, \bibinfo{person}{Iasonas Kokkinos}, \bibinfo{person}{Kevin
  Murphy}, {and} \bibinfo{person}{Alan~L Yuille}.}
  \bibinfo{year}{2017}\natexlab{}.
\newblock \showarticletitle{Deeplab: Semantic image segmentation with deep
  convolutional nets, atrous convolution, and fully connected crfs}.
\newblock \bibinfo{journal}{\emph{IEEE transactions on pattern analysis and
  machine intelligence}} \bibinfo{volume}{40}, \bibinfo{number}{4}
  (\bibinfo{year}{2017}), \bibinfo{pages}{834--848}.
\newblock


\bibitem[\protect\citeauthoryear{Chen, Li, Chen, and Gool}{Chen
  et~al\mbox{.}}{2019}]%
        {chen2019learning}
\bibfield{author}{\bibinfo{person}{Yuhua Chen}, \bibinfo{person}{Wen Li},
  \bibinfo{person}{Xiaoran Chen}, {and} \bibinfo{person}{Luc~Van Gool}.}
  \bibinfo{year}{2019}\natexlab{}.
\newblock \showarticletitle{Learning semantic segmentation from synthetic data:
  A geometrically guided input-output adaptation approach}. In
  \bibinfo{booktitle}{\emph{Proceedings of the IEEE Conference on Computer
  Vision and Pattern Recognition}}. \bibinfo{pages}{1841--1850}.
\newblock


\bibitem[\protect\citeauthoryear{Cordts, Omran, Ramos, Rehfeld, Enzweiler,
  Benenson, Franke, Roth, and Schiele}{Cordts et~al\mbox{.}}{2016}]%
        {cordts2016cityscapes}
\bibfield{author}{\bibinfo{person}{Marius Cordts}, \bibinfo{person}{Mohamed
  Omran}, \bibinfo{person}{Sebastian Ramos}, \bibinfo{person}{Timo Rehfeld},
  \bibinfo{person}{Markus Enzweiler}, \bibinfo{person}{Rodrigo Benenson},
  \bibinfo{person}{Uwe Franke}, \bibinfo{person}{Stefan Roth}, {and}
  \bibinfo{person}{Bernt Schiele}.} \bibinfo{year}{2016}\natexlab{}.
\newblock \showarticletitle{The cityscapes dataset for semantic urban scene
  understanding}. In \bibinfo{booktitle}{\emph{Proceedings of the IEEE
  Conference on Computer Vision and Pattern Recognition}}.
  \bibinfo{pages}{3213--3223}.
\newblock


\bibitem[\protect\citeauthoryear{Deng, Dong, Socher, Li, Li, and Fei-Fei}{Deng
  et~al\mbox{.}}{2009}]%
        {Imagenet2009}
\bibfield{author}{\bibinfo{person}{Jia Deng}, \bibinfo{person}{Wei Dong},
  \bibinfo{person}{Richard Socher}, \bibinfo{person}{Li-Jia Li},
  \bibinfo{person}{Kai Li}, {and} \bibinfo{person}{Li Fei-Fei}.}
  \bibinfo{year}{2009}\natexlab{}.
\newblock \showarticletitle{Imagenet: A large-scale hierarchical image
  database}. In \bibinfo{booktitle}{\emph{Proceedings of the IEEE Conference on
  Computer Vision and Pattern Recognition}}. \bibinfo{pages}{248--255}.
\newblock


\bibitem[\protect\citeauthoryear{Deng, Zheng, Ye, Kang, Yang, and Jiao}{Deng
  et~al\mbox{.}}{2018}]%
        {deng2018image}
\bibfield{author}{\bibinfo{person}{Weijian Deng}, \bibinfo{person}{Liang
  Zheng}, \bibinfo{person}{Qixiang Ye}, \bibinfo{person}{Guoliang Kang},
  \bibinfo{person}{Yi Yang}, {and} \bibinfo{person}{Jianbin Jiao}.}
  \bibinfo{year}{2018}\natexlab{}.
\newblock \showarticletitle{Image-image domain adaptation with preserved
  self-similarity and domain-dissimilarity for person re-identification}. In
  \bibinfo{booktitle}{\emph{Proceedings of the IEEE Conference on Computer
  Vision and Pattern Recognition}}. \bibinfo{pages}{994--1003}.
\newblock


\bibitem[\protect\citeauthoryear{Dosovitskiy, Ros, Codevilla, Lopez, and
  Koltun}{Dosovitskiy et~al\mbox{.}}{2017}]%
        {dosovitskiy2017carla}
\bibfield{author}{\bibinfo{person}{Alexey Dosovitskiy}, \bibinfo{person}{German
  Ros}, \bibinfo{person}{Felipe Codevilla}, \bibinfo{person}{Antonio Lopez},
  {and} \bibinfo{person}{Vladlen Koltun}.} \bibinfo{year}{2017}\natexlab{}.
\newblock \showarticletitle{CARLA: An open urban driving simulator}.
\newblock \bibinfo{journal}{\emph{arXiv preprint arXiv:1711.03938}}
  (\bibinfo{year}{2017}).
\newblock


\bibitem[\protect\citeauthoryear{Gaidon, Wang, Cabon, and Vig}{Gaidon
  et~al\mbox{.}}{2016}]%
        {gaidon2016virtual}
\bibfield{author}{\bibinfo{person}{Adrien Gaidon}, \bibinfo{person}{Qiao Wang},
  \bibinfo{person}{Yohann Cabon}, {and} \bibinfo{person}{Eleonora Vig}.}
  \bibinfo{year}{2016}\natexlab{}.
\newblock \showarticletitle{Virtual worlds as proxy for multi-object tracking
  analysis}. In \bibinfo{booktitle}{\emph{Proceedings of the IEEE Conference on
  Computer Vision and Pattern Recognition}}. \bibinfo{pages}{4340--4349}.
\newblock


\bibitem[\protect\citeauthoryear{Geiger, Lenz, Stiller, and Urtasun}{Geiger
  et~al\mbox{.}}{2013a}]%
        {geiger2013vision}
\bibfield{author}{\bibinfo{person}{Andreas Geiger}, \bibinfo{person}{Philip
  Lenz}, \bibinfo{person}{Christoph Stiller}, {and} \bibinfo{person}{Raquel
  Urtasun}.} \bibinfo{year}{2013}\natexlab{a}.
\newblock \showarticletitle{Vision meets robotics: The kitti dataset}.
\newblock \bibinfo{journal}{\emph{The International Journal of Robotics
  Research}} \bibinfo{volume}{32}, \bibinfo{number}{11} (\bibinfo{year}{2013}),
  \bibinfo{pages}{1231--1237}.
\newblock


\bibitem[\protect\citeauthoryear{Geiger, Lenz, Stiller, and Urtasun}{Geiger
  et~al\mbox{.}}{2013b}]%
        {Geiger2013IJRR}
\bibfield{author}{\bibinfo{person}{Andreas Geiger}, \bibinfo{person}{Philip
  Lenz}, \bibinfo{person}{Christoph Stiller}, {and} \bibinfo{person}{Raquel
  Urtasun}.} \bibinfo{year}{2013}\natexlab{b}.
\newblock \showarticletitle{Vision meets Robotics: The KITTI Dataset}.
\newblock \bibinfo{journal}{\emph{International Journal of Robotics Research}}
  (\bibinfo{year}{2013}).
\newblock


\bibitem[\protect\citeauthoryear{Geiger, Lenz, and Urtasun}{Geiger
  et~al\mbox{.}}{2012}]%
        {geiger2012we}
\bibfield{author}{\bibinfo{person}{Andreas Geiger}, \bibinfo{person}{Philip
  Lenz}, {and} \bibinfo{person}{Raquel Urtasun}.}
  \bibinfo{year}{2012}\natexlab{}.
\newblock \showarticletitle{Are we ready for autonomous driving? the kitti
  vision benchmark suite}. In \bibinfo{booktitle}{\emph{Proceedings of the IEEE
  Conference on Computer Vision and Pattern Recognition}}.
\newblock


\bibitem[\protect\citeauthoryear{Goodfellow, Pouget-Abadie, Mirza, Xu,
  Warde-Farley, Ozair, Courville, and Bengio}{Goodfellow et~al\mbox{.}}{2014}]%
        {NIPS2014_5423}
\bibfield{author}{\bibinfo{person}{Ian Goodfellow}, \bibinfo{person}{Jean
  Pouget-Abadie}, \bibinfo{person}{Mehdi Mirza}, \bibinfo{person}{Bing Xu},
  \bibinfo{person}{David Warde-Farley}, \bibinfo{person}{Sherjil Ozair},
  \bibinfo{person}{Aaron Courville}, {and} \bibinfo{person}{Yoshua Bengio}.}
  \bibinfo{year}{2014}\natexlab{}.
\newblock \showarticletitle{Generative Adversarial Nets}.
\newblock In \bibinfo{booktitle}{\emph{Advances in Neural Information
  Processing Systems}}. \bibinfo{pages}{2672--2680}.
\newblock


\bibitem[\protect\citeauthoryear{Gretton, Borgwardt, Rasch, Schoelkopf, and
  Smola}{Gretton et~al\mbox{.}}{2012}]%
        {Gretton2012}
\bibfield{author}{\bibinfo{person}{Arthur Gretton}, \bibinfo{person}{Karsten
  Borgwardt}, \bibinfo{person}{Malte~J Rasch}, \bibinfo{person}{Bernhard
  Schoelkopf}, {and} \bibinfo{person}{Alexander Smola}.}
  \bibinfo{year}{2012}\natexlab{}.
\newblock \showarticletitle{A Kernel Two-Sample Test}.
\newblock \bibinfo{journal}{\emph{Journal of Machine Learning Research}}
  \bibinfo{volume}{13} (\bibinfo{year}{2012}), \bibinfo{pages}{723--773}.
\newblock


\bibitem[\protect\citeauthoryear{Hattori, Naresh~Boddeti, Kitani, and
  Kanade}{Hattori et~al\mbox{.}}{2015}]%
        {hattori2015learning}
\bibfield{author}{\bibinfo{person}{Hironori Hattori}, \bibinfo{person}{Vishnu
  Naresh~Boddeti}, \bibinfo{person}{Kris~M Kitani}, {and}
  \bibinfo{person}{Takeo Kanade}.} \bibinfo{year}{2015}\natexlab{}.
\newblock \showarticletitle{Learning scene-specific pedestrian detectors
  without real data}. In \bibinfo{booktitle}{\emph{Proceedings of the IEEE
  Conference on Computer Vision and Pattern Recognition}}.
  \bibinfo{pages}{3819--3827}.
\newblock


\bibitem[\protect\citeauthoryear{Heusel, Ramsauer, Unterthiner, Nessler, and
  Hochreiter}{Heusel et~al\mbox{.}}{2017}]%
        {Heusel2017GANsTB}
\bibfield{author}{\bibinfo{person}{Martin Heusel}, \bibinfo{person}{Hubert
  Ramsauer}, \bibinfo{person}{Thomas Unterthiner}, \bibinfo{person}{Bernhard
  Nessler}, {and} \bibinfo{person}{Sepp Hochreiter}.}
  \bibinfo{year}{2017}\natexlab{}.
\newblock \showarticletitle{GANs Trained by a Two Time-Scale Update Rule
  Converge to a Local Nash Equilibrium}. In \bibinfo{booktitle}{\emph{Advances
  in Neural Information Processing Systems}}.
\newblock


\bibitem[\protect\citeauthoryear{Hoffman, Tzeng, Park, Zhu, Isola, Saenko,
  Efros, and Darrell}{Hoffman et~al\mbox{.}}{2017}]%
        {hoffman2017cycada}
\bibfield{author}{\bibinfo{person}{Judy Hoffman}, \bibinfo{person}{Eric Tzeng},
  \bibinfo{person}{Taesung Park}, \bibinfo{person}{Jun-Yan Zhu},
  \bibinfo{person}{Phillip Isola}, \bibinfo{person}{Kate Saenko},
  \bibinfo{person}{Alexei~A Efros}, {and} \bibinfo{person}{Trevor Darrell}.}
  \bibinfo{year}{2017}\natexlab{}.
\newblock \showarticletitle{Cycada: Cycle-consistent adversarial domain
  adaptation}.
\newblock \bibinfo{journal}{\emph{arXiv preprint arXiv:1711.03213}}
  (\bibinfo{year}{2017}).
\newblock


\bibitem[\protect\citeauthoryear{Hoffman, Wang, Yu, and Darrell}{Hoffman
  et~al\mbox{.}}{2016}]%
        {hoffman2016fcns}
\bibfield{author}{\bibinfo{person}{Judy Hoffman}, \bibinfo{person}{Dequan
  Wang}, \bibinfo{person}{Fisher Yu}, {and} \bibinfo{person}{Trevor Darrell}.}
  \bibinfo{year}{2016}\natexlab{}.
\newblock \showarticletitle{Fcns in the wild: Pixel-level adversarial and
  constraint-based adaptation}.
\newblock \bibinfo{journal}{\emph{arXiv preprint arXiv:1612.02649}}
  (\bibinfo{year}{2016}).
\newblock


\bibitem[\protect\citeauthoryear{Huang, Liu, Belongie, and Kautz}{Huang
  et~al\mbox{.}}{2018}]%
        {huang2018munit}
\bibfield{author}{\bibinfo{person}{Xun Huang}, \bibinfo{person}{Ming-Yu Liu},
  \bibinfo{person}{Serge Belongie}, {and} \bibinfo{person}{Jan Kautz}.}
  \bibinfo{year}{2018}\natexlab{}.
\newblock \showarticletitle{Multimodal Unsupervised Image-to-image
  Translation}. In \bibinfo{booktitle}{\emph{European Conference on Computer
  Vision}}.
\newblock


\bibitem[\protect\citeauthoryear{Isola, Zhu, Zhou, and Efros}{Isola
  et~al\mbox{.}}{2017}]%
        {Isola_2017_CVPR}
\bibfield{author}{\bibinfo{person}{Phillip Isola}, \bibinfo{person}{Jun-Yan
  Zhu}, \bibinfo{person}{Tinghui Zhou}, {and} \bibinfo{person}{Alexei~A.
  Efros}.} \bibinfo{year}{2017}\natexlab{}.
\newblock \showarticletitle{Image-To-Image Translation With Conditional
  Adversarial Networks}. In \bibinfo{booktitle}{\emph{Proceedings of the IEEE
  Conference on Computer Vision and Pattern Recognition}}.
\newblock


\bibitem[\protect\citeauthoryear{Juliani, Berges, Vckay, Gao, Henry, Mattar,
  and Lange}{Juliani et~al\mbox{.}}{2018}]%
        {DBLP:journals/corr/abs-1809-02627}
\bibfield{author}{\bibinfo{person}{Arthur Juliani},
  \bibinfo{person}{Vincent{-}Pierre Berges}, \bibinfo{person}{Esh Vckay},
  \bibinfo{person}{Yuan Gao}, \bibinfo{person}{Hunter Henry},
  \bibinfo{person}{Marwan Mattar}, {and} \bibinfo{person}{Danny Lange}.}
  \bibinfo{year}{2018}\natexlab{}.
\newblock \showarticletitle{Unity: {A} General Platform for Intelligent
  Agents}.
\newblock \bibinfo{journal}{\emph{CoRR}}  \bibinfo{volume}{abs/1809.02627}
  (\bibinfo{year}{2018}).
\newblock


\bibitem[\protect\citeauthoryear{Kar, Prakash, Liu, Cameracci, Yuan, Rusiniak,
  Acuna, Torralba, and Fidler}{Kar et~al\mbox{.}}{2019}]%
        {kar2019metasim}
\bibfield{author}{\bibinfo{person}{Amlan Kar}, \bibinfo{person}{Aayush
  Prakash}, \bibinfo{person}{Ming-Yu Liu}, \bibinfo{person}{Eric Cameracci},
  \bibinfo{person}{Justin Yuan}, \bibinfo{person}{Matt Rusiniak},
  \bibinfo{person}{David Acuna}, \bibinfo{person}{Antonio Torralba}, {and}
  \bibinfo{person}{Sanja Fidler}.} \bibinfo{year}{2019}\natexlab{}.
\newblock \showarticletitle{Meta-Sim: Learning to Generate Synthetic Datasets}.
  In \bibinfo{booktitle}{\emph{Proceedings of the IEEE International Conference
  on Computer Vision}}.
\newblock


\bibitem[\protect\citeauthoryear{Kolve, Mottaghi, Han, VanderBilt, Weihs,
  Herrasti, Gordon, Zhu, Gupta, and Farhadi}{Kolve et~al\mbox{.}}{2017}]%
        {ai2thor}
\bibfield{author}{\bibinfo{person}{Eric Kolve}, \bibinfo{person}{Roozbeh
  Mottaghi}, \bibinfo{person}{Winson Han}, \bibinfo{person}{Eli VanderBilt},
  \bibinfo{person}{Luca Weihs}, \bibinfo{person}{Alvaro Herrasti},
  \bibinfo{person}{Daniel Gordon}, \bibinfo{person}{Yuke Zhu},
  \bibinfo{person}{Abhinav Gupta}, {and} \bibinfo{person}{Ali Farhadi}.}
  \bibinfo{year}{2017}\natexlab{}.
\newblock \showarticletitle{{AI2-THOR: An Interactive 3D Environment for Visual
  AI}}.
\newblock \bibinfo{journal}{\emph{arXiv}} (\bibinfo{year}{2017}).
\newblock


\bibitem[\protect\citeauthoryear{Li, Liang, Jia, and Xing}{Li
  et~al\mbox{.}}{2018}]%
        {li2018semantic}
\bibfield{author}{\bibinfo{person}{Peilun Li}, \bibinfo{person}{Xiaodan Liang},
  \bibinfo{person}{Daoyuan Jia}, {and} \bibinfo{person}{Eric~P Xing}.}
  \bibinfo{year}{2018}\natexlab{}.
\newblock \showarticletitle{Semantic-aware grad-gan for virtual-to-real urban
  scene adaption}.
\newblock \bibinfo{journal}{\emph{arXiv preprint arXiv:1801.01726}}
  (\bibinfo{year}{2018}).
\newblock


\bibitem[\protect\citeauthoryear{Li, Fang, Yang, Wang, Lu, and Yang}{Li
  et~al\mbox{.}}{2017}]%
        {NIPS2017_6642}
\bibfield{author}{\bibinfo{person}{Yijun Li}, \bibinfo{person}{Chen Fang},
  \bibinfo{person}{Jimei Yang}, \bibinfo{person}{Zhaowen Wang},
  \bibinfo{person}{Xin Lu}, {and} \bibinfo{person}{Ming-Hsuan Yang}.}
  \bibinfo{year}{2017}\natexlab{}.
\newblock \showarticletitle{Universal Style Transfer via Feature Transforms}.
\newblock In \bibinfo{booktitle}{\emph{Advances in Neural Information
  Processing Systems}}, \bibfield{editor}{\bibinfo{person}{I.~Guyon},
  \bibinfo{person}{U.~V. Luxburg}, \bibinfo{person}{S.~Bengio},
  \bibinfo{person}{H.~Wallach}, \bibinfo{person}{R.~Fergus},
  \bibinfo{person}{S.~Vishwanathan}, {and} \bibinfo{person}{R.~Garnett}}
  (Eds.). \bibinfo{pages}{386--396}.
\newblock


\bibitem[\protect\citeauthoryear{Long, Shelhamer, and Darrell}{Long
  et~al\mbox{.}}{2015}]%
        {long2015fully}
\bibfield{author}{\bibinfo{person}{Jonathan Long}, \bibinfo{person}{Evan
  Shelhamer}, {and} \bibinfo{person}{Trevor Darrell}.}
  \bibinfo{year}{2015}\natexlab{}.
\newblock \showarticletitle{Fully convolutional networks for semantic
  segmentation}. In \bibinfo{booktitle}{\emph{Proceedings of the IEEE
  Conference on Computer Vision and Pattern Recognition}}.
  \bibinfo{pages}{3431--3440}.
\newblock


\bibitem[\protect\citeauthoryear{Luo, Zheng, Guan, Yu, and Yang}{Luo
  et~al\mbox{.}}{2019}]%
        {luo2019taking}
\bibfield{author}{\bibinfo{person}{Yawei Luo}, \bibinfo{person}{Liang Zheng},
  \bibinfo{person}{Tao Guan}, \bibinfo{person}{Junqing Yu}, {and}
  \bibinfo{person}{Yi Yang}.} \bibinfo{year}{2019}\natexlab{}.
\newblock \showarticletitle{Taking a closer look at domain shift:
  Category-level adversaries for semantics consistent domain adaptation}. In
  \bibinfo{booktitle}{\emph{Proceedings of the IEEE Conference on Computer
  Vision and Pattern Recognition}}. \bibinfo{pages}{2507--2516}.
\newblock


\bibitem[\protect\citeauthoryear{Pepik, Stark, Gehler, and Schiele}{Pepik
  et~al\mbox{.}}{2012}]%
        {pepik2012teaching}
\bibfield{author}{\bibinfo{person}{Bojan Pepik}, \bibinfo{person}{Michael
  Stark}, \bibinfo{person}{Peter Gehler}, {and} \bibinfo{person}{Bernt
  Schiele}.} \bibinfo{year}{2012}\natexlab{}.
\newblock \showarticletitle{Teaching 3d geometry to deformable part models}. In
  \bibinfo{booktitle}{\emph{Proceedings of the IEEE Conference on Computer
  Vision and Pattern Recognition}}.
\newblock


\bibitem[\protect\citeauthoryear{Prakash, Boochoon, Brophy, Acuna, Cameracci,
  State, Shapira, and Birchfield}{Prakash et~al\mbox{.}}{2019}]%
        {prakash2019structured}
\bibfield{author}{\bibinfo{person}{Aayush Prakash}, \bibinfo{person}{Shaad
  Boochoon}, \bibinfo{person}{Mark Brophy}, \bibinfo{person}{David Acuna},
  \bibinfo{person}{Eric Cameracci}, \bibinfo{person}{Gavriel State},
  \bibinfo{person}{Omer Shapira}, {and} \bibinfo{person}{Stan Birchfield}.}
  \bibinfo{year}{2019}\natexlab{}.
\newblock \showarticletitle{Structured domain randomization: Bridging the
  reality gap by context-aware synthetic data}. In
  \bibinfo{booktitle}{\emph{Proceedings of the IEEE International Conference on
  Robotics and Automation}}. \bibinfo{pages}{7249--7255}.
\newblock


\bibitem[\protect\citeauthoryear{Richter, Vineet, Roth, and Koltun}{Richter
  et~al\mbox{.}}{2016}]%
        {Richter_2016_ECCV}
\bibfield{author}{\bibinfo{person}{Stephan~R. Richter}, \bibinfo{person}{Vibhav
  Vineet}, \bibinfo{person}{Stefan Roth}, {and} \bibinfo{person}{Vladlen
  Koltun}.} \bibinfo{year}{2016}\natexlab{}.
\newblock \showarticletitle{Playing for Data: {G}round Truth from Computer
  Games}. In \bibinfo{booktitle}{\emph{European Conference on Computer
  Vision}}.
\newblock


\bibitem[\protect\citeauthoryear{Ros, Sellart, Materzynska, Vazquez, and
  Lopez}{Ros et~al\mbox{.}}{2016}]%
        {ros2016synthia}
\bibfield{author}{\bibinfo{person}{German Ros}, \bibinfo{person}{Laura
  Sellart}, \bibinfo{person}{Joanna Materzynska}, \bibinfo{person}{David
  Vazquez}, {and} \bibinfo{person}{Antonio~M Lopez}.}
  \bibinfo{year}{2016}\natexlab{}.
\newblock \showarticletitle{The synthia dataset: A large collection of
  synthetic images for semantic segmentation of urban scenes}. In
  \bibinfo{booktitle}{\emph{Proceedings of the IEEE Conference on Computer
  Vision and Pattern Recognition}}. \bibinfo{pages}{3234--3243}.
\newblock


\bibitem[\protect\citeauthoryear{Ruiz, Schulter, and Chandraker}{Ruiz
  et~al\mbox{.}}{2018}]%
        {ruiz2018learning}
\bibfield{author}{\bibinfo{person}{Nataniel Ruiz}, \bibinfo{person}{Samuel
  Schulter}, {and} \bibinfo{person}{Manmohan Chandraker}.}
  \bibinfo{year}{2018}\natexlab{}.
\newblock \showarticletitle{Learning to simulate}.
\newblock \bibinfo{journal}{\emph{arXiv preprint arXiv:1810.02513}}
  (\bibinfo{year}{2018}).
\newblock


\bibitem[\protect\citeauthoryear{Sakaridis, Dai, and Van~Gool}{Sakaridis
  et~al\mbox{.}}{2018}]%
        {sakaridis2018semantic}
\bibfield{author}{\bibinfo{person}{Christos Sakaridis},
  \bibinfo{person}{Dengxin Dai}, {and} \bibinfo{person}{Luc Van~Gool}.}
  \bibinfo{year}{2018}\natexlab{}.
\newblock \showarticletitle{Semantic foggy scene understanding with synthetic
  data}.
\newblock \bibinfo{journal}{\emph{International Journal of Computer Vision}}
  \bibinfo{volume}{126}, \bibinfo{number}{9} (\bibinfo{year}{2018}),
  \bibinfo{pages}{973--992}.
\newblock


\bibitem[\protect\citeauthoryear{Satkin, Lin, and Hebert}{Satkin
  et~al\mbox{.}}{2012}]%
        {satkin2012data}
\bibfield{author}{\bibinfo{person}{Scott Satkin}, \bibinfo{person}{Jason Lin},
  {and} \bibinfo{person}{Martial Hebert}.} \bibinfo{year}{2012}\natexlab{}.
\newblock \showarticletitle{Data-driven scene understanding from 3D models}. In
  \bibinfo{booktitle}{\emph{European Conference on Computer Vision}}.
\newblock


\bibitem[\protect\citeauthoryear{Simonyan and Zisserman}{Simonyan and
  Zisserman}{2014}]%
        {simonyan2014very}
\bibfield{author}{\bibinfo{person}{Karen Simonyan} {and}
  \bibinfo{person}{Andrew Zisserman}.} \bibinfo{year}{2014}\natexlab{}.
\newblock \showarticletitle{Very deep convolutional networks for large-scale
  image recognition}.
\newblock \bibinfo{journal}{\emph{arXiv preprint arXiv:1409.1556}}
  (\bibinfo{year}{2014}).
\newblock


\bibitem[\protect\citeauthoryear{Sun and Zheng}{Sun and Zheng}{2019}]%
        {sun2019dissecting}
\bibfield{author}{\bibinfo{person}{Xiaoxiao Sun} {and} \bibinfo{person}{Liang
  Zheng}.} \bibinfo{year}{2019}\natexlab{}.
\newblock \showarticletitle{Dissecting Person Re-identification from the
  Viewpoint of Viewpoint}. In \bibinfo{booktitle}{\emph{Proceedings of the IEEE
  Conference on Computer Vision and Pattern Recognition}}.
\newblock


\bibitem[\protect\citeauthoryear{Tobin, Fong, Ray, Schneider, Zaremba, and
  Abbeel}{Tobin et~al\mbox{.}}{2017}]%
        {tobin2017domain}
\bibfield{author}{\bibinfo{person}{Josh Tobin}, \bibinfo{person}{Rachel Fong},
  \bibinfo{person}{Alex Ray}, \bibinfo{person}{Jonas Schneider},
  \bibinfo{person}{Wojciech Zaremba}, {and} \bibinfo{person}{Pieter Abbeel}.}
  \bibinfo{year}{2017}\natexlab{}.
\newblock \showarticletitle{Domain randomization for transferring deep neural
  networks from simulation to the real world}. In
  \bibinfo{booktitle}{\emph{Proceedings of the IEEE International Conference on
  Intelligent Robots and Systems}}.
\newblock


\bibitem[\protect\citeauthoryear{Tremblay, Prakash, Acuna, Brophy, Jampani,
  Anil, To, Cameracci, Boochoon, and Birchfield}{Tremblay
  et~al\mbox{.}}{2018}]%
        {Tremblay2018TrainingDN}
\bibfield{author}{\bibinfo{person}{Jonathan Tremblay}, \bibinfo{person}{Aayush
  Prakash}, \bibinfo{person}{David Acuna}, \bibinfo{person}{Mark Brophy},
  \bibinfo{person}{Varun Jampani}, \bibinfo{person}{Cem Anil},
  \bibinfo{person}{Thang To}, \bibinfo{person}{Eric Cameracci},
  \bibinfo{person}{Shaad Boochoon}, {and} \bibinfo{person}{Stanley~T.
  Birchfield}.} \bibinfo{year}{2018}\natexlab{}.
\newblock \showarticletitle{Training Deep Networks with Synthetic Data:
  Bridging the Reality Gap by Domain Randomization}.
\newblock \bibinfo{journal}{\emph{Proceedings of the IEEE Conference on
  Computer Vision and Pattern Recognition Workshops}} (\bibinfo{year}{2018}).
\newblock


\bibitem[\protect\citeauthoryear{Tsai, Hung, Schulter, Sohn, Yang, and
  Chandraker}{Tsai et~al\mbox{.}}{2018}]%
        {tsai2018learning}
\bibfield{author}{\bibinfo{person}{Yi-Hsuan Tsai}, \bibinfo{person}{Wei-Chih
  Hung}, \bibinfo{person}{Samuel Schulter}, \bibinfo{person}{Kihyuk Sohn},
  \bibinfo{person}{Ming-Hsuan Yang}, {and} \bibinfo{person}{Manmohan
  Chandraker}.} \bibinfo{year}{2018}\natexlab{}.
\newblock \showarticletitle{Learning to adapt structured output space for
  semantic segmentation}. In \bibinfo{booktitle}{\emph{Proceedings of the IEEE
  Conference on Computer Vision and Pattern Recognition}}.
  \bibinfo{pages}{7472--7481}.
\newblock


\bibitem[\protect\citeauthoryear{Williams}{Williams}{1992}]%
        {williams1992simple}
\bibfield{author}{\bibinfo{person}{Ronald~J Williams}.}
  \bibinfo{year}{1992}\natexlab{}.
\newblock \showarticletitle{Simple statistical gradient-following algorithms
  for connectionist reinforcement learning}.
\newblock \bibinfo{journal}{\emph{Machine learning}} \bibinfo{volume}{8},
  \bibinfo{number}{3-4} (\bibinfo{year}{1992}), \bibinfo{pages}{229--256}.
\newblock


\bibitem[\protect\citeauthoryear{Wright}{Wright}{2015}]%
        {wright2015coordinate}
\bibfield{author}{\bibinfo{person}{Stephen~J Wright}.}
  \bibinfo{year}{2015}\natexlab{}.
\newblock \showarticletitle{Coordinate descent algorithms}.
\newblock \bibinfo{journal}{\emph{Mathematical Programming}}
  \bibinfo{volume}{151}, \bibinfo{number}{1} (\bibinfo{year}{2015}),
  \bibinfo{pages}{3--34}.
\newblock


\bibitem[\protect\citeauthoryear{Wu, Wu, Gkioxari, and Tian}{Wu
  et~al\mbox{.}}{2018}]%
        {Wu2018BuildingGA}
\bibfield{author}{\bibinfo{person}{Yi Wu}, \bibinfo{person}{Yuxin Wu},
  \bibinfo{person}{Georgia Gkioxari}, {and} \bibinfo{person}{Yuandong Tian}.}
  \bibinfo{year}{2018}\natexlab{}.
\newblock \showarticletitle{Building Generalizable Agents with a Realistic and
  Rich 3D Environment}.
\newblock \bibinfo{journal}{\emph{ArXiv}}  \bibinfo{volume}{abs/1801.02209}
  (\bibinfo{year}{2018}).
\newblock


\bibitem[\protect\citeauthoryear{Yao, Zheng, Yang, Naphade, and Gedeon}{Yao
  et~al\mbox{.}}{2019}]%
        {yao2019simulating}
\bibfield{author}{\bibinfo{person}{Yue Yao}, \bibinfo{person}{Liang Zheng},
  \bibinfo{person}{Xiaodong Yang}, \bibinfo{person}{Milind Naphade}, {and}
  \bibinfo{person}{Tom Gedeon}.} \bibinfo{year}{2019}\natexlab{}.
\newblock \showarticletitle{Simulating Content Consistent Vehicle Datasets with
  Attribute Descent}.
\newblock \bibinfo{journal}{\emph{arXiv preprint arXiv:1912.08855}}
  (\bibinfo{year}{2019}).
\newblock


\bibitem[\protect\citeauthoryear{Zhang, Qiu, Yao, Liu, and Mei}{Zhang
  et~al\mbox{.}}{2018}]%
        {zhang2018fully}
\bibfield{author}{\bibinfo{person}{Yiheng Zhang}, \bibinfo{person}{Zhaofan
  Qiu}, \bibinfo{person}{Ting Yao}, \bibinfo{person}{Dong Liu}, {and}
  \bibinfo{person}{Tao Mei}.} \bibinfo{year}{2018}\natexlab{}.
\newblock \showarticletitle{Fully convolutional adaptation networks for
  semantic segmentation}. In \bibinfo{booktitle}{\emph{Proceedings of the IEEE
  Conference on Computer Vision and Pattern Recognition}}.
  \bibinfo{pages}{6810--6818}.
\newblock


\bibitem[\protect\citeauthoryear{Zhao, Shi, Qi, Wang, and Jia}{Zhao
  et~al\mbox{.}}{2017}]%
        {Zhao_2017_CVPR}
\bibfield{author}{\bibinfo{person}{Hengshuang Zhao}, \bibinfo{person}{Jianping
  Shi}, \bibinfo{person}{Xiaojuan Qi}, \bibinfo{person}{Xiaogang Wang}, {and}
  \bibinfo{person}{Jiaya Jia}.} \bibinfo{year}{2017}\natexlab{}.
\newblock \showarticletitle{Pyramid Scene Parsing Network}. In
  \bibinfo{booktitle}{\emph{Proceedings of the IEEE Conference on Computer
  Vision and Pattern Recognition}}.
\newblock


\bibitem[\protect\citeauthoryear{Zheng, Shen, Tian, Wang, Wang, and Tian}{Zheng
  et~al\mbox{.}}{2015}]%
        {zheng2015scalable}
\bibfield{author}{\bibinfo{person}{Liang Zheng}, \bibinfo{person}{Liyue Shen},
  \bibinfo{person}{Lu Tian}, \bibinfo{person}{Shengjin Wang},
  \bibinfo{person}{Jingdong Wang}, {and} \bibinfo{person}{Qi Tian}.}
  \bibinfo{year}{2015}\natexlab{}.
\newblock \showarticletitle{Scalable person re-identification: A benchmark}. In
  \bibinfo{booktitle}{\emph{Proceedings of the IEEE International Conference on
  Computer Vision}}. \bibinfo{pages}{1116--1124}.
\newblock


\bibitem[\protect\citeauthoryear{Zhu, Park, Isola, and Efros}{Zhu
  et~al\mbox{.}}{2017}]%
        {Zhu_2017_ICCV}
\bibfield{author}{\bibinfo{person}{Jun-Yan Zhu}, \bibinfo{person}{Taesung
  Park}, \bibinfo{person}{Phillip Isola}, {and} \bibinfo{person}{Alexei~A.
  Efros}.} \bibinfo{year}{2017}\natexlab{}.
\newblock \showarticletitle{Unpaired Image-To-Image Translation Using
  Cycle-Consistent Adversarial Networks}. In
  \bibinfo{booktitle}{\emph{Proceedings of the IEEE International Conference on
  Computer Vision}}.
\newblock


\bibitem[\protect\citeauthoryear{Zou, Yu, Kumar, and Wang}{Zou
  et~al\mbox{.}}{2018}]%
        {zou2018domain}
\bibfield{author}{\bibinfo{person}{Yang Zou}, \bibinfo{person}{Zhiding Yu},
  \bibinfo{person}{BVK Kumar}, {and} \bibinfo{person}{Jinsong Wang}.}
  \bibinfo{year}{2018}\natexlab{}.
\newblock \showarticletitle{Domain adaptation for semantic segmentation via
  class-balanced self-training}.
\newblock \bibinfo{journal}{\emph{arXiv preprint arXiv:1810.07911}}
  (\bibinfo{year}{2018}).
\newblock


\end{thebibliography}

\appendix

\begin{figure}[h]
    \centering
    \includegraphics[scale=0.3]{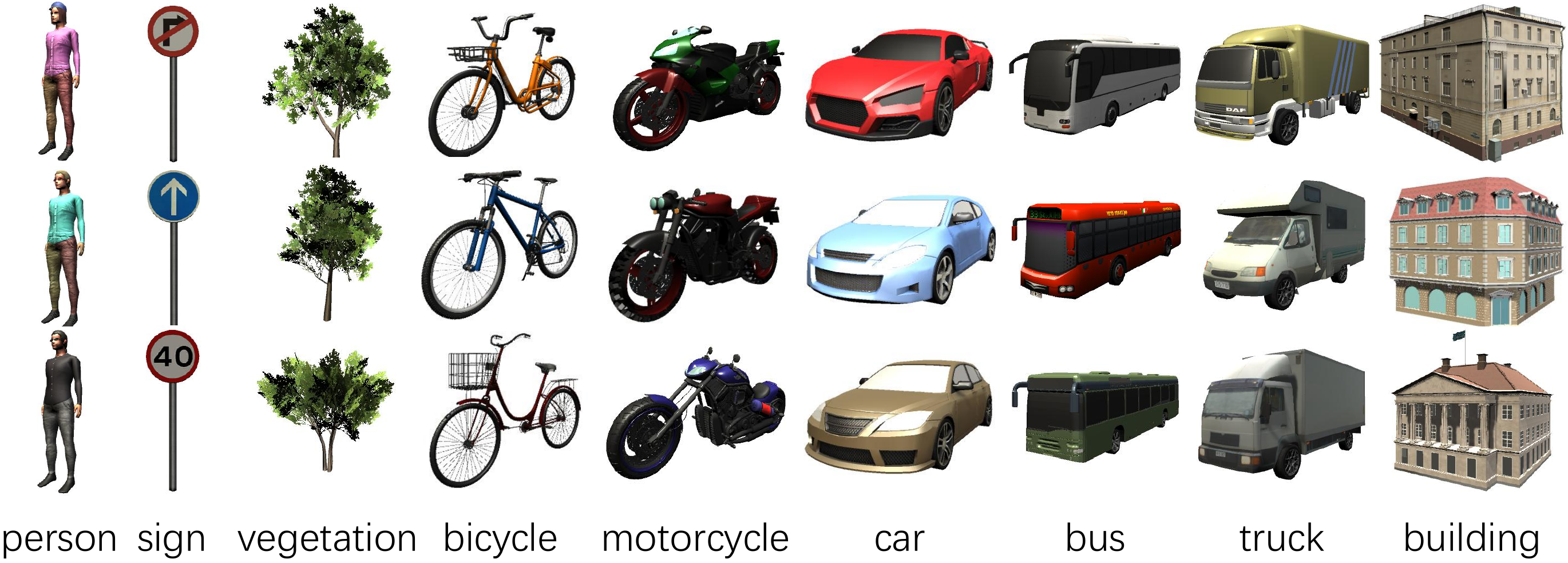}
    \caption{Example assets in SceneX. Assets are available for all 19 classes in Cityscapes \cite{cordts2016cityscapes}. 3D Models in each category have various appearances to simulate diverse scenes.}
    \label{fig:unity_assets}
\end{figure}

\begin{figure}[h]
\centering
\includegraphics[scale=0.5]{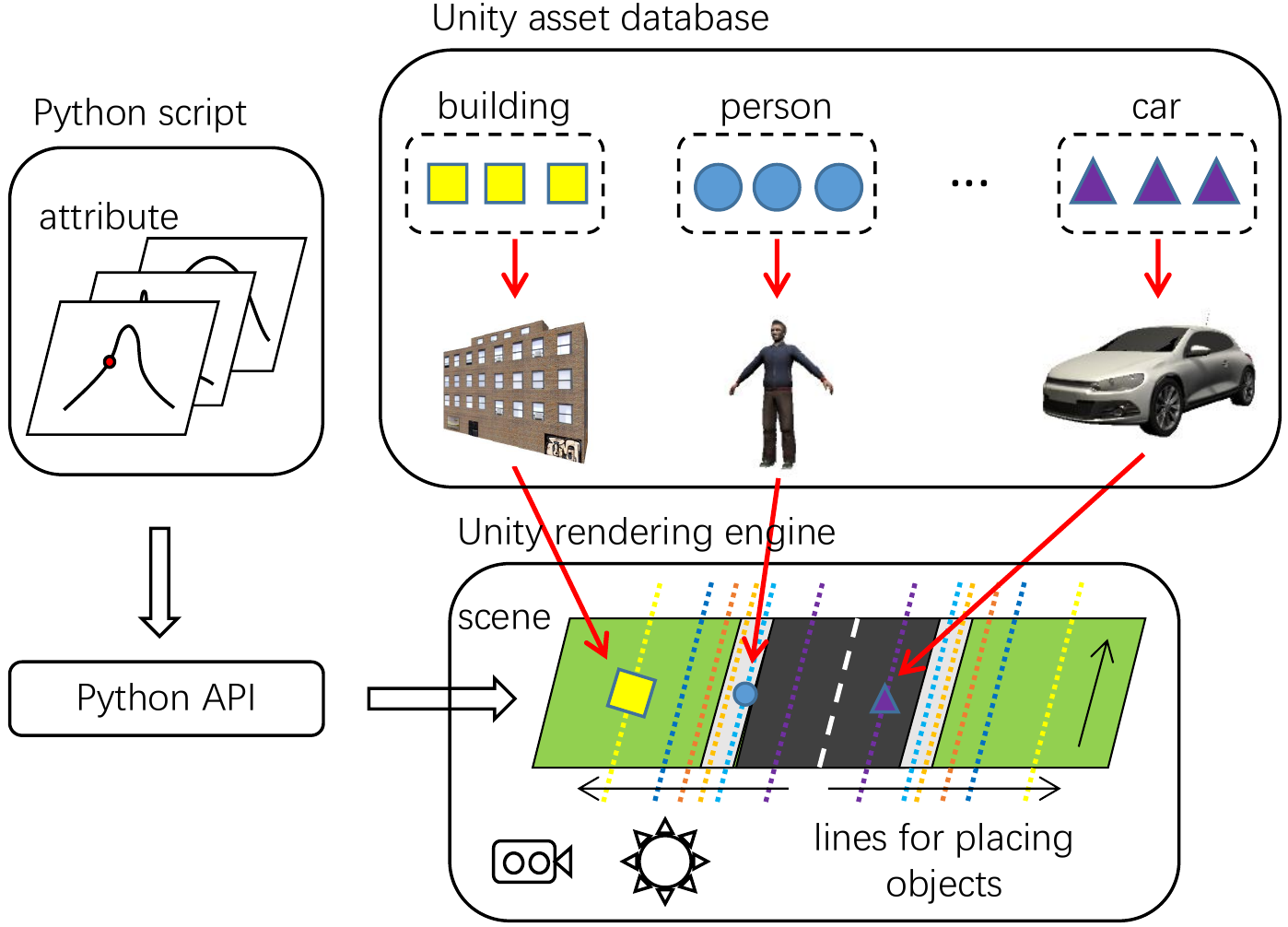}
\caption{Details of SceneX. It is composed of Unity asset database, Unity rendering engine and a Python API. The Unity rendering engine is featured by a ``line-based'' design. Through Python programming, we can easily control the scene structure by modifying the global attributes.}
\label{fig:scenex}
\end{figure}

\section{Details of SceneX}

\subsection{3D Scene Classes and Assets}
To perform segmentation on a rich range of objects, we have collected a large number of 3D assets for the engine. SceneX contains 19 classes of virtual objects, \emph{i.e.,} car, pedestrian, building and \emph{etc}, which is compatible with Cityscapes \cite{cordts2016cityscapes}.
Specifically, 
SceneX contains 200 pedestrian models, 195 cars, 28 buses and 39 trucks from existing model repositories \cite{sun2019dissecting,yao2019simulating}, and makes necessary modifications so that they are compatible with our engine.
Besides, we collect 106 buildings, 18 bicycles, and 19 trees, among others. There are also 14 sky box models to simulate different weather conditions.
Some sample object models are shown in Fig. \ref{fig:unity_assets}.

\subsection{Engine design}
As illustrated in Fig. \ref{fig:scenex}, our system is mainly composed of Unity asset database, Unity rendering engine and a Python API.
The Unity database contains 19 classes of 3D assets that have various appearances, and it is extendable. The Unity rendering engine defines the scene structure and is able to change the environment variables.
The scene structure is featured by a ``line-based'' design.
After the scene is constructed, the camera moves and captures the scene, outputting a set of synthetic images as well as corresponding ground truth segmentation labels, which are automatically generated through rendering buffer. The Python API ensures us to control the scene structure by modifying the global attributes within SceneX, through Python programming.

The ``line-based'' design enables us to control the scene structure with few parameters. Specifically, as shown in Fig. \ref{fig:scenex}, to the center of a scene is a road map, around which are a group of lines. These lines are designed to place objects on.
Same types of objects (\emph{e.g.,} bicycles) are placed along a line parallel to the road, and they share the distance with the road.
Objects on the same line are tied together, which means changing the position of an object equals to moving all the objects placed by the same magnitude. This object placement strategy not only allows us to easily adjust the distance between objects and the road, but also enables precise object density changes by modifying the interval between objects on the same line.

Thus, the scene generation process can be viewed as a sequential process. Firstly, the positions of the lines are determined, and the types of objects to be placed on them are determined as well. Then, objects such as buildings, persons and cars are randomly picked from the Unity asset database, and placed onto their corresponding lines. After all the lines are filled with objects, the illumination changes and the camera moves to capture the scene. After that, the scene is destroyed and another scene is constructed. This scene generation process repeats several times, such that the generated dataset consists of various scenes.

\textbf{Label acquisition.} The advantage of data synthesis is that labels can be obtained freely. 
Given an image of the scene, pixel-level ground truths can be obtained through the rendering buffer. 

\subsection{Attribute design}
There 23 global attributes within SceneX, and they are classified into three groups, \emph{i.e.}, 8 for environment, 7 for object position (\emph{i.e.,} line position) and 8 for density. The details of these attributes are listed as follow.

\textbf{Environment variables.}
\begin{itemize}
    \item [-] Illumination intensity, that changes the brightness of the virtual environment.
    \item [-] Illumination angle x, that changes the rotation angle along x axis for illumination.
    \item [-] Illumination angle y, that changes the rotation angle along y axis for illumination.
    \item [-] Camera position probability, that changes the probability of camera to be on left or right side on the road.
    \item [-] Camera position x, that changes the camera position along x axis.
    \item [-] Camera position y, that changes the camera position along y axis.
    \item [-] Camera rotation x, that changes the camera rotation angle along x axis.
    \item [-] Camera rotation y, that changes the camera rotation angle along y axis.
\end{itemize}

\textbf{Position variables for context splines.}
\begin{itemize}
    \item [-] Building position, that changes the parallel distance between the building (and train) and the road.
    \item [-] Fence position, that changes the parallel distance between the fence (and wall) and the road.
    \item [-] Tree position, that changes the parallel distance between the tree and the road.
    \item [-] Bicycle position, that changes the parallel distance between the bicycle (and motorcycle) and the road.
    \item [-] Person position, that changes the parallel distance between the person (and rider) and the road.
    \item [-] Pole position, that changes the parallel distance between the pole (and traffic sign) and the road.
    \item [-] Car position, that changes the parallel distance between the car (and bus, truck) and the road.
\end{itemize}

\textbf{Density variables for context splines.}
\begin{itemize}
    \item [-] Building probability, that changes the occurrence probability for building (and train).
    \item [-] Building interval, that changes the interval for buildings.
    \item [-] Fence interval, that changes the interval for fences (and walls).
    \item [-] Tree interval, that changes the interval for trees.
    \item [-] Bicycle interval, that changes the interval for bicycles (and motorcycles).
    \item [-] Person interval, that changes the interval for pedestrians (and riders).
    \item [-] Pole interval, that changes the interval for poles (and traffic signs).
    \item [-] Car interval, that changes the interval for cars (and buses and trucks).
\end{itemize}

\section{Details of attribute training}

\subsection{Cityscapes as target dataset}
When using Cityscapes as the target dataset, we simulate synthetic images at resolution of $640 \times 320$ at training stage, and $1024 \times 512$ at testing stage. We manually permutate the 23 global attributes into three groups, \emph{i.e.,} 7 for object locations, 8 for object densities and 8 for environments. Furthermore, we split the 8 object density variables into four groups, with two attributes in each group.
Thus, the 23 attributes are split into six groups, and they are optimized using SDR in groups.

During attribute training, each group of attributes are optimized for up to 50 updates of the policy network.
For each update, one dataset is generated with a size of 180 images, and thus an accuracy score is calculated to update the policy network.

\subsection{CamVid as target dataset}
When using CamVid as the target dataset, we simulate synthetic images at resolution of $480 \times 360$ at both training and testing stages.
The 8 object location variables are split into four groups, with two attributes in each group respectively.
Thus, the 23 attributes in this case are split into six groups. And we simulate 180 images in each optimizing step like Cityscapes.


\begin{figure*}[t]
    \vspace{50pt}
    \centering
    \includegraphics[scale=0.53]{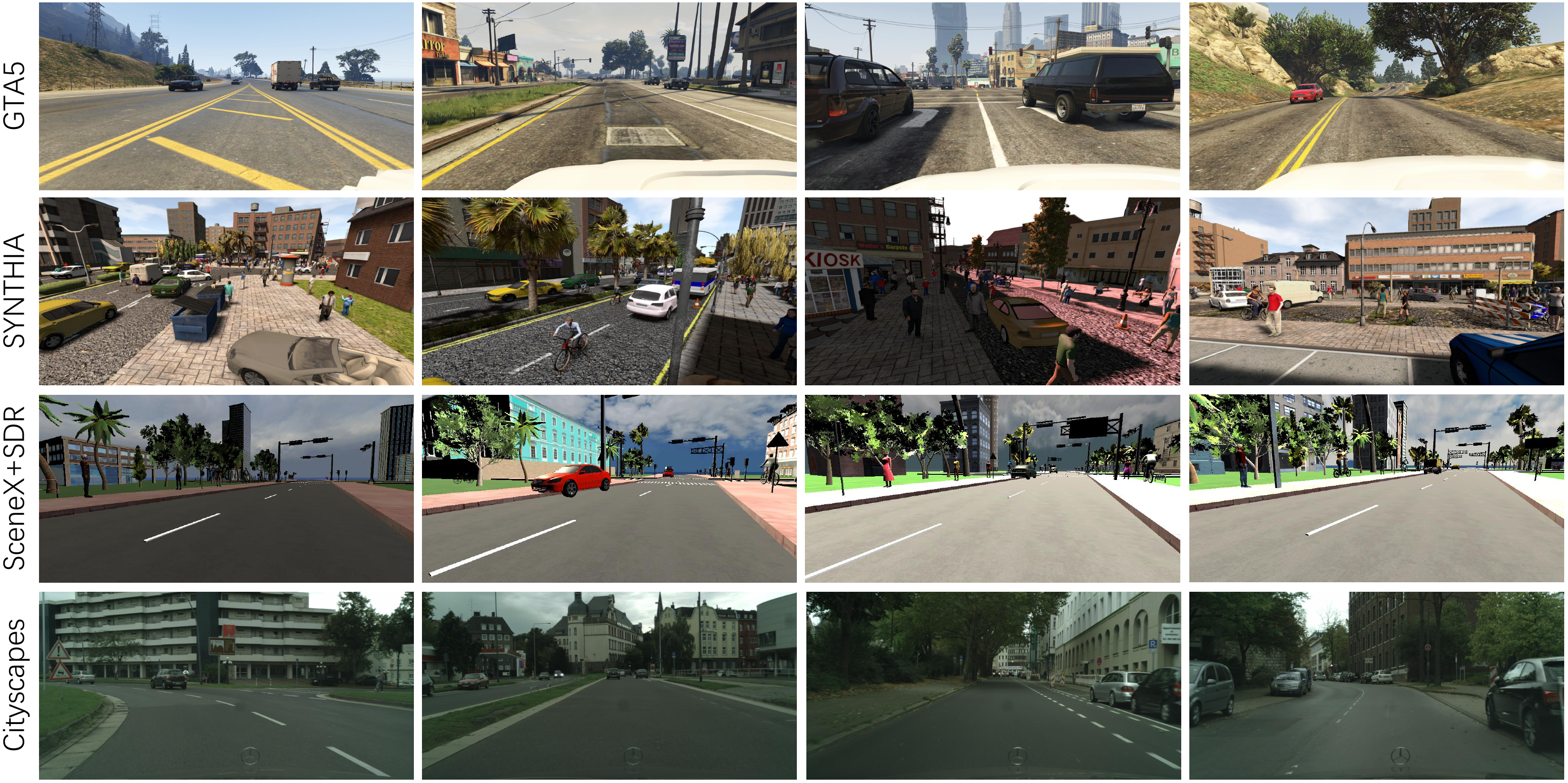}
    \caption{Sample images of existing synthetic datasets (a) GTA5 and (b) SYNTHIA and (c) simulated dataset using SDR within SceneX (SceneX+SDR) as well as the target real-world dataset (d) Cityscapes.}
    \label{fig:synthetic_data_city}
\end{figure*}

\begin{figure*}[t]
    \centering
    \vspace{50pt}
    \includegraphics[scale=0.62]{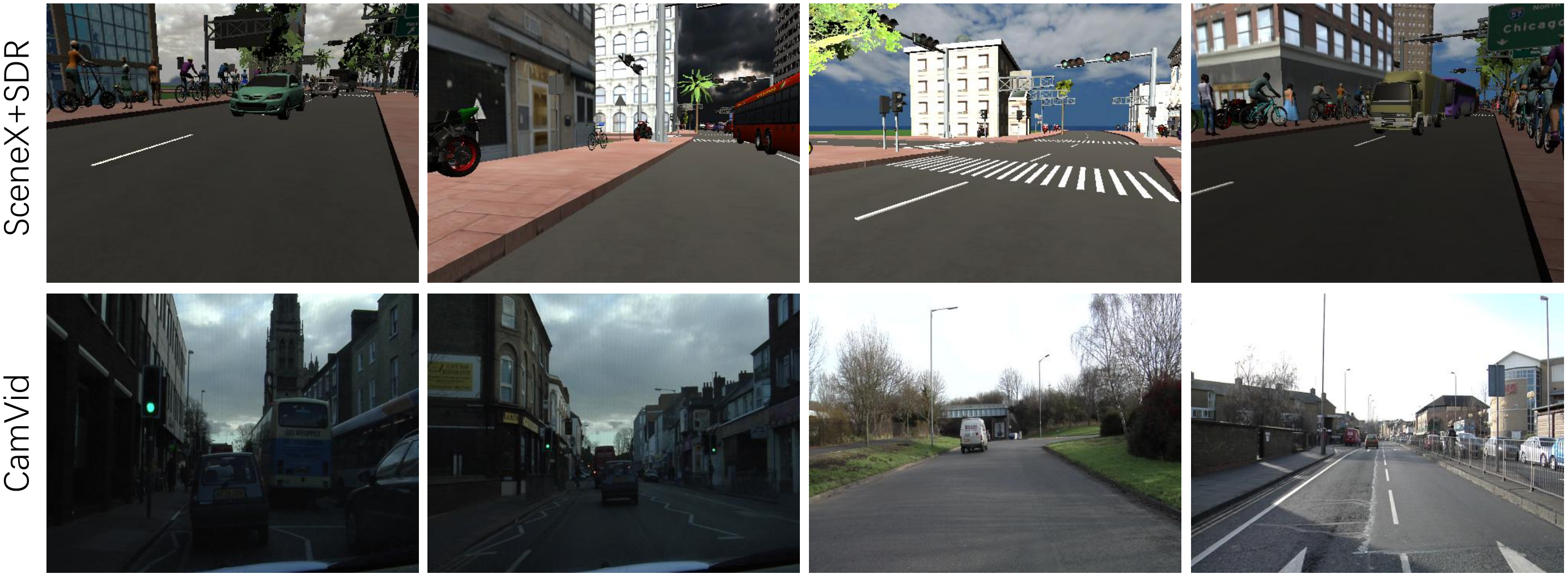}
    \caption{Sample images of (top) simulated images using SDR within SceneX (SceneX+SDR) as well as (bottom) the target dataset CamVid.}
    \label{fig:synthetic_data_cam}
\end{figure*}


\end{document}